\documentclass{article}
\pdfoutput=1
\usepackage{pdfx}

\usepackage{iclr2022_conference,times}

\usepackage[utf8]{inputenc} %
\usepackage[T1]{fontenc}    %
\usepackage{hyperref}       %
\usepackage{url}            %
\usepackage{booktabs}       %
\usepackage{amsfonts}       %
\usepackage{nicefrac}       %
\usepackage{microtype}      %
\usepackage{xcolor}         %
\usepackage{wrapfig}
\usepackage{amsmath}
\usepackage{cleveref}
\usepackage{subcaption}
\usepackage{csquotes}
\usepackage{graphicx}
\usepackage{multirow}
\usepackage{enumitem}
\usepackage{xspace}

\newcommand{\paragraphsq}[1]{\vspace{-5pt}\paragraph{#1}}
\usepackage{caption}
\newcommand{\tabbelowsqueeze}{\vspace{-9pt}}

\definecolor{bleudefrance}{rgb}{0.19, 0.55, 0.91}
\definecolor{cornflowerblue}{rgb}{0.39, 0.58, 0.93}
\definecolor{darkcerulean}{rgb}{0.03, 0.27, 0.49}
\definecolor{deepskyblue}{rgb}{0.0, 0.75, 1.0}

\definecolor{deepgreen}{rgb}{0.0, 0.6, 0.0}

\newcommand{\eat}[1]{}

\makeatletter
\g@addto@macro\small{%
  \setlength\abovedisplayskip{-5pt}
  \setlength\abovedisplayshortskip{-5pt}
  \setlength\belowdisplayshortskip{-5pt}
  \setlength\belowdisplayskip{-5pt}
}
\makeatother

\newcommand{\norm}[1]{\left\lVert#1\right\rVert}

\newcommand{\wavvec}{W2V2\xspace}
\newcommand{\wavvecsize}[1]{\wavvec-#1\xspace}
\newcommand{\finalmodel}{SEW\xspace} 
\newcommand{\finalmodelD}{\finalmodel-D\xspace} 
\newcommand{\finalmodelsize}[1]{\finalmodel-#1\xspace}
\newcommand{\finalmodelDsize}[1]{\finalmodelD-#1\xspace}
\newcommand{\finalmodelfull}{Squeezed and Efficient Wav2vec} 

\newlength\savewidth
\newcommand{\tablestyle}[2]{\setlength{\tabcolsep}{#1}\renewcommand{\arraystretch}{#2}\centering\footnotesize}
\makeatletter\renewcommand\paragraph{\@startsection{paragraph}{4}{\z@}
  {.5em \@plus1ex \@minus.2ex}{-.5em}{\normalfont\normalsize\bfseries}}\makeatother
\newcommand{\mypm}[1]{\color{gray}{\tiny{$\pm$#1}}}

\newcommand*{\shortautoref}[1]{%
  \begingroup
    \def\sectionautorefname{Sec.}%
    \def\subsectionautorefname{Subsec.}%
    \def\figureautorefname{Fig.}%
    \def\algorithmautorefname{Alg.}%
    \def\listingautorefname{List.}%
    \autoref{#1}%
  \endgroup
}

\usepackage{amsmath,amsfonts,bm}

\def\eqref#1{equation~\ref{#1}}

\def\1{\bm{1}}

\def\rmC{{\mathbf{C}}}

\def\rmQ{{\mathbf{Q}}}

\def\rmX{{\mathbf{X}}}

\def\rmZ{{\mathbf{Z}}}

\def\va{{\bm{a}}}
\def\vb{{\bm{b}}}
\def\vc{{\bm{c}}}

\def\ve{{\bm{e}}}

\def\vm{{\bm{m}}}

\def\vq{{\bm{q}}}

\def\vx{{\bm{x}}}

\def\vz{{\bm{z}}}

\def\mA{{\bm{A}}}

\def\mC{{\bm{C}}}

\def\mK{{\bm{K}}}

\def\mO{{\bm{O}}}
\def\mP{{\bm{P}}}
\def\mQ{{\bm{Q}}}

\def\mV{{\bm{V}}}
\def\mW{{\bm{W}}}

\DeclareMathAlphabet{\mathsfit}{\encodingdefault}{\sfdefault}{m}{sl}
\SetMathAlphabet{\mathsfit}{bold}{\encodingdefault}{\sfdefault}{bx}{n}

\def\gL{{\mathcal{L}}}

\newcommand{\E}{\mathbb{E}}

\newcommand{\R}{\mathbb{R}}

\DeclareMathOperator*{\argmax}{arg\,max}

\title{Performance-Efficiency Trade-offs \\ in Unsupervised Pre-training \\ for Speech Recognition}

\author{
  Felix Wu$\phantom{}^{\dagger}$  \hspace{0pt} 
  Kwangyoun Kim$\phantom{}^{\dagger}$  \hspace{0pt} 
  Jing Pan$\phantom{}^{\dagger}$  \hspace{0pt}
  Kyu Han$\phantom{}^{\dagger}$  \hspace{0pt} 
  Kilian Q. Weinberger$\phantom{}^{\dagger\ddagger}$ \hspace{0pt} Yoav Artzi$\phantom{}^{\dagger\ddagger}$ \\[3pt]
  $\phantom{}^{\dagger}$ASAPP Inc.\hspace{15pt}$\phantom{}^{\ddagger}$Cornell University \\
  \texttt{\{fwu, kkim, jpan, khan, kweinberger, yoav\}@asapp.com} \\
}

\iclrfinalcopy %

\begin{document}

\maketitle

\begin{abstract}
  This paper is a study of performance-efficiency trade-offs in pre-trained models for automatic speech recognition (ASR). 
We focus on wav2vec 2.0, and formalize several architecture designs that influence both the model performance and its efficiency. 
Putting together all our observations, we introduce \textit{SEW (Squeezed and Efficient Wav2vec)}, a pre-trained model architecture with significant improvements along both performance and efficiency dimensions across a variety of training setups. 
For example, under the 100h-960h semi-supervised setup on LibriSpeech, SEW achieves a 1.9x inference speedup compared to wav2vec 2.0, with a 13.5\% relative reduction in word error rate. With a similar inference time, SEW reduces word error rate by 25--50\% across different model sizes.

\end{abstract}

\section{Introduction} \label{sec:intro}

\begin{wrapfigure}{r}{0.4\textwidth}
\vspace{-20pt}
\includegraphics[width=\linewidth]{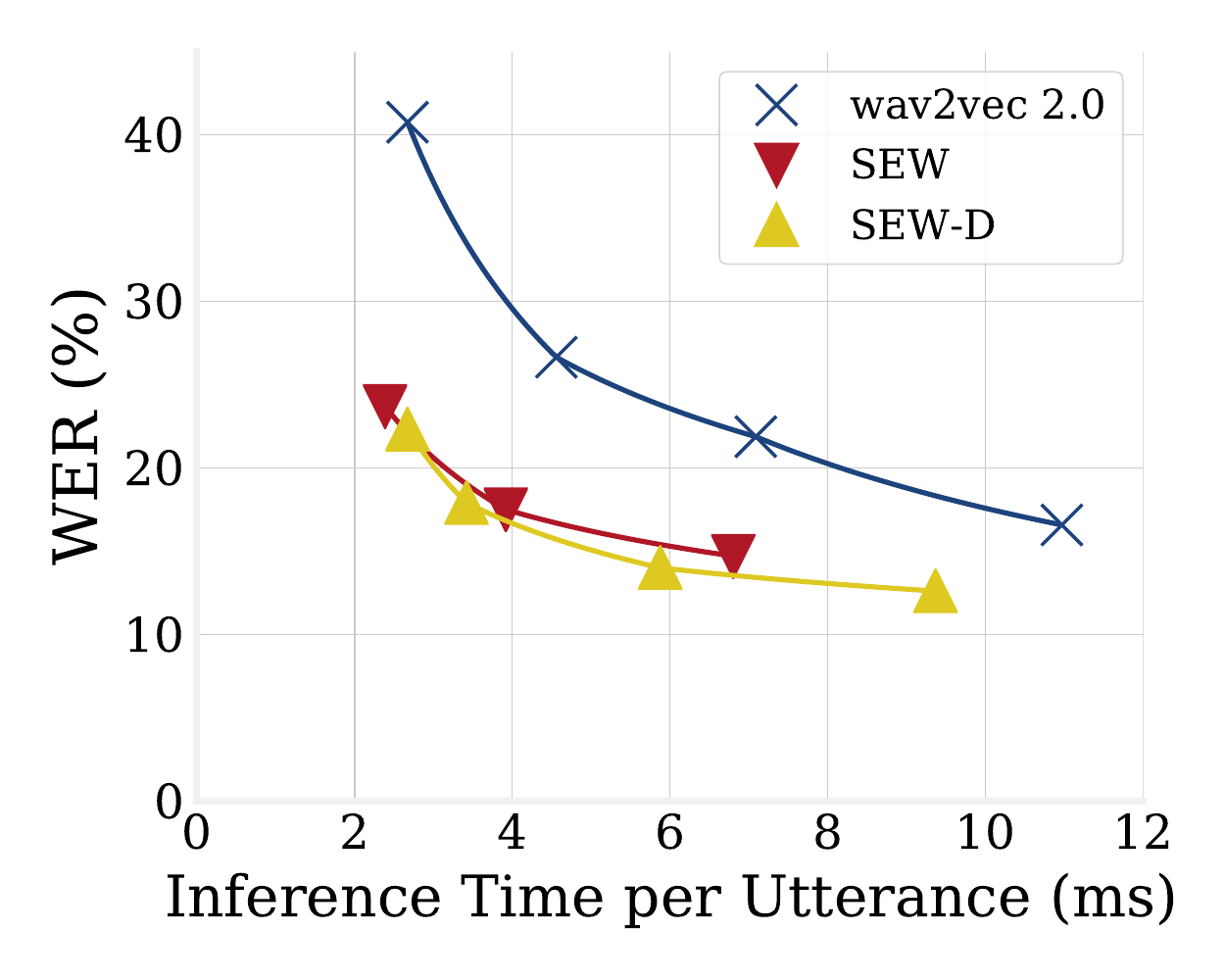} 
\caption{Word error rate (WER) and average utterance inference time on LibriSpeech (dev-other) of wav2vec 2.0 and our \finalmodel and \finalmodelD models fine-tuned with 100h labeled data for 100K updates.}
\label{fig:w2v2_vs_sew}
\vspace{-5pt}
\end{wrapfigure}

Recently, there is significant interest in self-supervised pre-training using unlabeled audio data to learn versatile feature representations, that are subsequently fine-tuned on task-specific annotated audio~\citep{Zhang2020PushingTL,Wang2021LargeScaleSA,Xu2020SelftrainingAP,Pepino2021EmotionRF}. 
This follows similar trends in natural language processing~\cite[NLP;][]{devlin2018bert,liu2019roberta,He2020DeBERTaDB} and computer vision~\cite[CV;][]{he2019moco,Chen2020SimCLR,grill2020bootstrap}. 
Maybe the most prominent example of this class of models is wav2vec 2.0~\citep[W2V2;][]{baevski2020wav2vec2}, which achieves competitive word error rate (WER) following fine-tuning on only ten minutes of transcribed (labeled) data, when prior supervised approaches often require nearly a thousand hours. 
If recent developments in NLP and CV are any indication, the importance of such pre-trained audio models that are fine-tuned on expert tasks will only increase. 
Indeed, W2V2 has already been studied with focus on the impact of pre-training data~\citep{Conneau2020UnsupervisedCR,Hsu2021RobustW2}, pre-training task~\citep{hsu2020hubert}, or combination with pseudo labelling~\citep{Xu2020SelftrainingAP,Zhang2020PushingTL}.

In this paper, we study W2V2's model design, and possible trade-offs between its components. 
Our focus is on efficiency for practical applications, rather than extending the model. 
As W2V2-type models become increasingly common, understanding their efficiency trade-offs is critical for transferring their benefits from the lab to the real world, where any increase in efficiency can substantially reduce the inference costs and energy footprints across a plethora of real world applications.

We study several aspects of the W2V2 model. 
We focus on automatic speech recognition (ASR), while retaining the standard pre-training and few-sample fine-tuning setup.\footnote{While pre-training time can be considered a secondary metric for efficiency, it is not our primary goal.} 
First, we study how the temporal resolution of the network trades-off performance and efficiency, and show that using different resolutions for computing pre-trained representations and ASR decoding significantly reduces inference time, while retaining similar performance. 
Second, we propose an efficient family of waveform feature extractors, which achieves similar performance with half the inference time as the original W2V2 extractor.
Finally, we study the impact of shifting model expressivity between different parts of the network. We observe that it is better to assign more parameters to later parts in the pre-trained network, compared to increasing capacity closer to the input waveform. We also see that increasing the expressivity of the pre-training predictor heads increases performance, while not influencing downstream-task computation as these heads are discarded.

We combine our observations to propose two models: \finalmodel\ (\finalmodelfull) and \finalmodel-D (\finalmodel\ with disentangled attention~\citep{He2020DeBERTaDB}).
We pre-train \finalmodel\ and \finalmodel-D on 960 hours of unlabelled audio from the LibriSpeech dataset~\citep{Panayotov2015LibrispeechAA}, and fine-tune with multiple ASR tasks. 
\finalmodel\ yields a significantly better performance-efficiency trade-off than the original W2V2.
For example, with 100h labeled data, compared to a \wavvecsize{tiny} model, 
\finalmodel\ reduces the LibriSpeech test-clean WER from 22.8\% to 10.6\% while being slightly faster, even outperforming a larger \wavvec model with 12.8\% WER. 
Compared to the official \wavvecsize{large} release, our best \finalmodelDsize{base+} achieves 2.7$\times$ and 3.2$\times$ speed-ups for inference and pre-training with comparable WER using half of the number of parameters.
Compared to \wavvecsize{base}, our \finalmodelDsize{mid} achieves 1.9$\times$ inference speed-up with a 13.5\% relative reduction in WER.
\shortautoref{fig:w2v2_vs_sew} shows the performance-efficiency trade-offs with various model size.
\finalmodel-D outperforms W2V2 in most pre-training settings, when experimenting with LibriSpeech~\citep{Panayotov2015LibrispeechAA}, Ted-lium 3~\citep{Hernandez2018TEDLIUM3T}, VoxPopuli~\citep{Wang2021VoxPopuliAL}, and Switchboard~\citep{switchboard} datasets. 
Pre-trained models and code are available at \url{https://github.com/asappresearch/sew}.

\section{Related Work}\label{sec:related_works}

\paragraphsq{Unsupervised Audio Representation Learning}
Contrastive predictive coding (CPC) is a general unsupervised learning method for speech, vision, text, and reinforcement learning~\citep{Oord2018RepresentationLW}.
When applied to speech, it uses past audio to predict the future audio, similar to language modeling~\citep{mikolov2010recurrent,dauphin2017language,kaplan2020scaling} but with contrastive loss.
Wav2vec~\citep{Schneider2019wav2vecUP} further improves the CPC model architecture design and focuses on unsupervised pre-training for end-to-end automatic speech recognition. 
Roughly speaking, wav2vec includes a feature extractor that generates a sequence of vectors from raw waveform audio, and a context network that encodes the features from the recent past to predict the features in the immediate future. 
This context network is only used  to learn useful feature representations, and is typically discarded after pre-training. 
Recently, \citet{Baevski2020vqwav2vecSL} introduced vq-wav2vec and a combination of vq-wav2vec with a discrete BERT-like model~\citep{Devlin2019BERT,Baevski2019EffectivenessOS}. 
W2V2~\citep{baevski2020wav2vec2} combines vq-wav2vec and the BERT-like model into an end-to-end setup, where the BERT portion functions as the context network, but not discarded. 
More recently, \citet{hsu2020hubert} propose HuBERT and show that W2V2 can be pre-trained with clustered targets instead of contrastive objectives.
Besides ASR-focused works, there is significant interest in learning representations for other speech tasks~\citep{synnaeve2016temporal,chung2018unsupervised,chuang2019speechbert,song2019speech}, music~\citep{yang2021deeper,zhao2021musicoder}, and general audio~\citep{saeed2021contrastive,gong2021psla,niizumi2021byol,wang2021multimodal}.

\paragraphsq{End-to-end Automatic Speech Recognition (ASR)}
As large datasets and fast compute become available, end-to-end ASR models~\citep{amodei2016deep,Zhang2020PushingTL} increasingly achieve state-of-the-art results, outperforming HMM-DNN hybrid systems~\citep{abdel2012applying,hinton2012deep}.
End-to-end ASR models can be roughly categorized into three main types: connectionist temporal classification~\citep[CTC;][]{graves2013speech}, RNN transducers~\citep[RNN-T;][]{graves2012sequence,Han2020ContextNetIC,Gulati2020ConformerCT}, and sequence-to-sequence  (a.k.a. Listen, Attend and Spell models)~\citep[Seq2seq;][]{chan2016listen,dong2018speech,watanabe2018espnet}.
CTC models are extremely fast for batch decoding; RNN-T variants are often used in real-time systems; Seq2seq models are more popular in offline settings.
Recently, and following success on NLP tasks, there is a transition in speech processing towards the Transformer architecture~\citep{Vaswani2017AttentionIA,dong2018speech} and its variants~\citep{Zhang2020TransformerTA,baevski2020wav2vec2,Gulati2020ConformerCT,Zhang2020PushingTL,yeh2019transformer}.
\section{Technical Background: Wav2Vec 2.0 (W2V2)}\label{sec:w2v2}

W2V2 is made of a waveform feature extractor that generates a sequence of continuous feature vectors, each encoding a small segment of audio, and a context network that maps these vectors to context-dependent representations. 

\begin{wrapfigure}{r}{0.43\textwidth}
\vspace{-20pt}
\includegraphics[width=\linewidth]{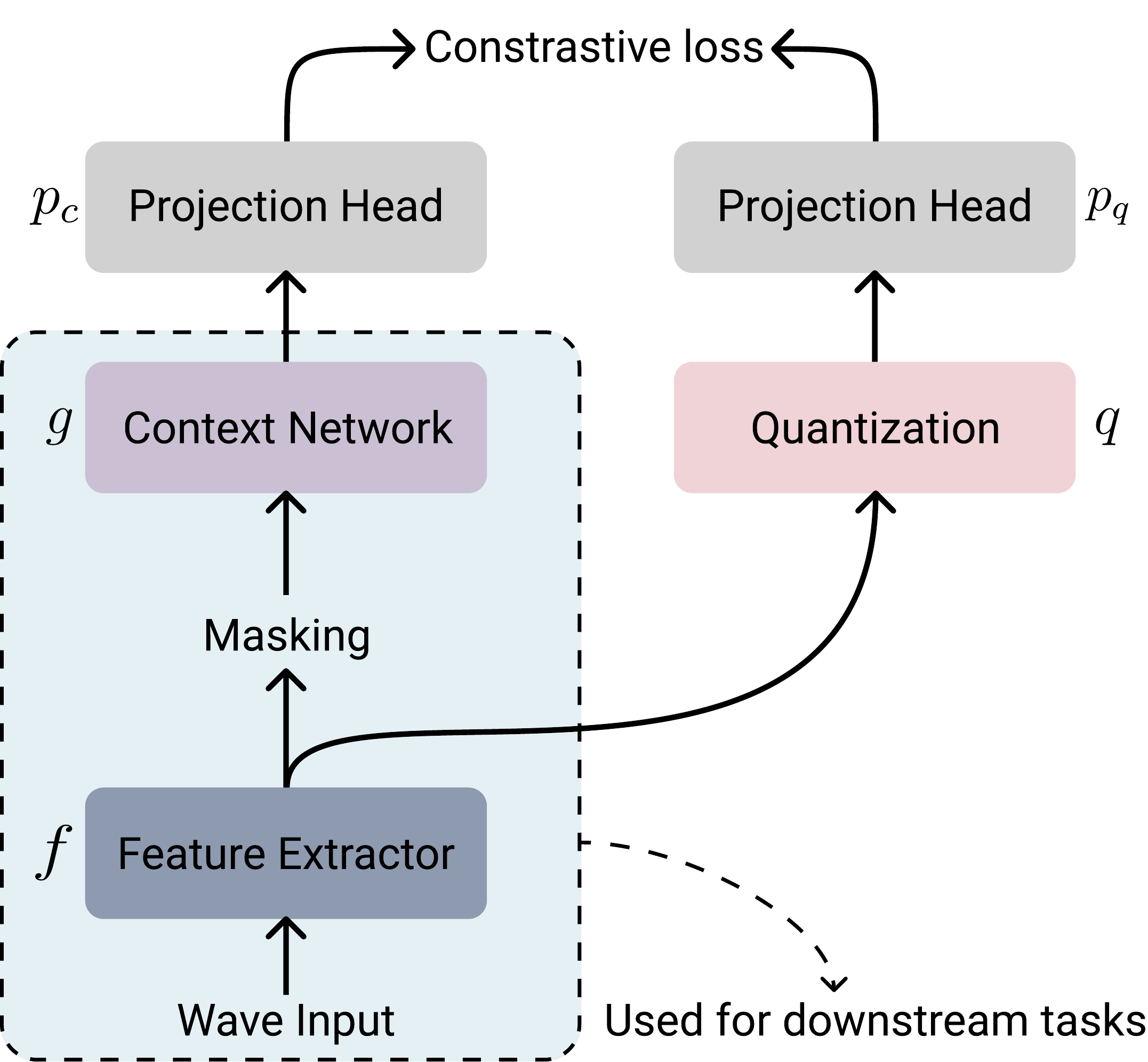} 
\caption{Wav2vec 2.0 framework.}
\label{fig:w2v2_framework}
\vspace{-20pt}
\end{wrapfigure}

During pre-training, some of the features are masked out, and are not seen by the context network. 
In parallel, the pre-masking features are discretized as prediction targets.
The context network aims to discriminate the discretized version of the original features at the masked positions from a pool of negative samples using the InfoNCE loss~\citep{Oord2018RepresentationLW}.

\mbox{\shortautoref{fig:w2v2_framework}} shows the \wavvec framework, including (a) a feature extractor, (b) a context network, (c) an optional quantization module, and (d) two projection heads.

\paragraphsq{Wave Feature Extractor (WFE)}
The wave feature extractor $f(\cdot)$ encodes and downsamples the raw waveform audio inputs $\rmX = (\vx_1, ..., \vx_{T_\text{input}}) \in \R^{T_\text{input} \times d_\text{input}}$ ($d_\text{input} = 1$ for single-channel audio) into an array of feature vectors $\rmZ= f(\rmX) = (\vz_1, ..., \vz_T) \in \R^{T \times \R^{d_\text{feat}}}$. 
For example, \wavvec maps 16KHz audio sequences to 50Hz frames using a convolutional WFE with receptive field size of 400 and stride size of 320. 
Each feature vector encodes the raw signals within a 25ms ($= 1000 / 16000 * 400$) window with a stride size 20ms ($= 1000 / 16000 * 320$). 
The reduced sequence length is $T = \frac{T_\text{input} - 400}{320} + 1 = \frac{T_\text{input} - 80}{320}$.

\paragraphsq{Context Network}
The context network $g(\cdot)$ follows a similar principle as masked language models in NLP (e.g.,  BERT~\citep{devlin2018bert} or RoBERTa~\citep{liu2019roberta}).
During pre-training, each $z_t$ is masked and replaced with a trainable mask vector $\vm$ with a predefined probability $p$. To illustrate, $\rmZ = (\vz_1, \vz_2, \vz_3, \vz_4, \vz_5, \vz_6, ..., \vz_T)$ can become $\rmZ' = (\vz_1, \vm, \vm, \vz_4, \vm, \vz_6, ..., \vz_T)$.
The context network maps this masked sequence to a sequence of contextual representations $\rmC = g(\rmZ') = (\vc_1, ..., \vc_T) \in \R^{T \times d_\text{feat}}$ to incorporate context information. Even if $\vz_t$ is masked and replaced with $\vm$, we anticipate that $\vc_t$  can recover the information in $\vz_t$ because it contains information from surrounding, not-masked input vectors. 
The context network is usually implemented with a Transformer architecture~\citep{Vaswani2017AttentionIA,Gulati2020ConformerCT}. 

\paragraphsq{Quantization Module}
The quatization module $q(\cdot)$ maps each unmasked vector $\vz_t$ into a quantized form $\vq_t = q(\vz_t) \in \R^{d_\text{feat}}$ for each masked position $t$.
Quantized $\vq_t$'s are the prediction targets. 
The quantization module is based on Gumbel softmax with straight-through estimator~\citep{gumbel1954statistical,jang2016categorical,maddison2014sampling}.
There are $G$ codebooks and each codebook has $V$ entries, giving  $G \times V$ vectors $\ve_{g,v} \in \R^{\frac{d_\text{feat}}{G}}$ where $g \in \{1, ..., G\}, v \in \{1, ..., V\}$. 
For each group $g$, the probability of assigning $\vz_t$ to the $v$-th entry is $p_{g,v} = \frac{\exp(\mW^g_v \cdot \vz_t / \tau_Q)}{\sum_{v'=1}^V \exp(\mW^g_{v'} \cdot \vz_t / \tau_Q)}$, where $\mW^g \in \R^{V \times d_\text{feat}}$ is a trainable matrix and $\tau_Q$ is the quantization temperature. For each group $g$, $\vz_t$ is assigned to the $v_g^*$-th entry where $v_g^* = \argmax_{v} p_{g,v}$. The corresponding embedding vectors $(\ve_{1, v_1^*}, ..., \ve_{G, v_G^*}$) are concatenated to a single vector $\vq_t \in \R^{d_\text{feat}}$, and constitutes a quantized feature sequence $\rmQ = (\vq_1, ..., \vq_T) \in \R^{T \times d_\text{feat}}$.

\paragraphsq{Projection Heads}
Two linear projection heads $p_c(\cdot)$ and $p_q(\cdot)$ reduce the dimensionality of $\rmC$ and $\rmQ$.
For a $\vz_t$  that is masked and replaced with $\vm$, we want $p_c(\vc_t) \in \R^{d_\text{proj}}$ to be similar to $p_q(\vq_t) \in \R^{d_\text{proj}}$.
\citet{baevski2020wav2vec2} do not separate between $p_c$ and $g$ or $p_q$ and $q$ in their original notations. However, we keep the distinctions, as they serve different roles and are discarded before downstream fine-tuning. 

\paragraphsq{Pre-training Objective}
\wavvec combines contrastive and diversity losses in the pre-training loss:
\begin{equation}
\gL = \gL_m + \alpha \gL_d\;\;.
\end{equation}
The goal of the contrastive loss $\gL_m$ is to make the projected outputs $p_c(\vc_t)$ close to $p_q(\vq_t)$ and far away from any other $p_q(\vq_{t'})$, where $\vz_t$ is masked and $t'$ is any other position in the same sequence.
\wavvec uses an InfoNCE loss~\citep{Oord2018RepresentationLW}:

\begin{small}
\begin{equation}
    \gL_m = \E_{t \text{\ is masked}} \left[ -\log \frac{\exp(\mathrm{sim}(p_c(\vc_t), p_q(\vq_t)) / \kappa)}{\sum_{\vq_{t'} \in \mathbb{Q}} \exp(\mathrm{sim}(p_c(\vc_t), p_q(\vq_{t'})) / \kappa)} \right],
\end{equation}
\end{small}

\noindent
with $\mathrm{sim}(\va, \vb) = \frac{\va^\top \vb}{\norm{\va} \norm{\vb}}$, $\mathbb{Q}$ is a set containing the positive sample $\vq_t$ and $K$ negative samples, and $\kappa$ is the temperature. The expectation is computed  over masked positions only.
The diversity loss $\gL_d$ prevents the quantization module from collapsing to a trivial mapping (e.g., by collapsing all inputs to a single discrete code). It encourages the quantization probability $p_{g, v}$ to be evenly distributed:

\begin{small}
\begin{equation}
    \gL_d = \E_{t} \left[ 1 - \frac{1}{GV} \sum_{g=1}^G \exp \left( - \sum_{v=1}^V p_{g,v} \log p_{g,v} \right) \right].
\end{equation}
\end{small}

\section{Exploring Model Design Trade-offs} \label{sec:method}

\subsection{Experimental Setup} \label{subsec:exp_setup}

We use the official \wavvec implementation in fairseq~\citep{Ott2019fairseqAF},  with the hyper-parameters of \wavvecsize{base}~\citep{baevski2020wav2vec2}. 
We describe key hyper-parameters; the linked  configuration files provide the full details.

\paragraphsq{Pre-training}
We use the LibriSpeech (CC BY 4.0)~\citep{Panayotov2015LibrispeechAA} 960h training data for unsupervised pre-training, leaving 1\% as validation set for pre-training.
We use the same hyperparameters as \wavvecsize{base}\footnote{\href{https://github.com/pytorch/fairseq/blob/master/examples/wav2vec/config/pretraining/wav2vec2_base_librispeech.yaml}{https://github.com/pytorch/fairseq/blob/master/examples/wav2vec/config/pretraining/wav2vec2\_base\_librispeech.yaml}} (\shortautoref{app:sec:exp_setup_details}).
To speed up and reduce the cost of our experiments, we pre-train all models for 100K updates similar to \citet{hsu2020hubert}.
All experiments use an AWS p3.16xlarge instance with 8 NVIDIA V100 GPUs and 64 Intel Xeon 2.30GHz CPU cores.
Because \citet{baevski2020wav2vec2} use 64 GPUs, we set gradient accumulation steps to 8 to simulate their 64-GPU pre-training with 8 GPUs.

\paragraphsq{Fine-tuning}
We add a linear classifier to the top of the context network and fine-tune the model using a CTC objective on LibriSpeech train-clean 100h set for 80K updates using the same set of hyper-parameters as \wavvecsize{base} (\shortautoref{app:sec:exp_setup_details}).\footnote{\href{https://github.com/pytorch/fairseq/blob/master/examples/wav2vec/config/finetuning/base_100h.yaml}{https://github.com/pytorch/fairseq/blob/master/examples/wav2vec/config/finetuning/base\_100h.yaml}}

\paragraphsq{Evaluation}
We use CTC greedy decoding~\citep{Graves2006ctc} for all experiments because it is faster than Viterbi decoding~\citep{Viterbi1967ErrorBF} and we do not find any WER differences between the two using baseline \wavvec models (\shortautoref{app:sec:ctc_decoding}). 
We use LibriSpeech dev-other for validation, and hold out test-clean and test-other as test sets. 
We consider three metrics to evaluate model efficiency and  performance: pre-training time, inference time, and WER (word error rate).
All evaluation is done on a NVIDIA V100 GPU with FP32 operations, unless specified otherwise.
When decoding with a language model (LM), we use the official 4-gram LM\footnote{\href{https://www.openslr.org/resources/11/4-gram.arpa.gz}{https://www.openslr.org/resources/11/4-gram.arpa.gz}} and wav2letter~\citep{Collobert2016Wav2LetterAE} decoder\footnote{\href{https://github.com/flashlight/wav2letter/tree/v0.2/bindings/python}{https://github.com/flashlight/wav2letter/tree/v0.2/bindings/python}} with the default LM weight 2, word score -1, and beam size 50.
Reducing the inference time with LM is an important direction for future work, as the wav2letter decoder is the bottleneck and is at least 3$\times$ slower than \wavvecsize{base} (\shortautoref{app:sec:ctc_decoding}).

\subsection{Depth vs. Width} \label{subsec:depth_vs_width}

The smallest \wavvec model is \wavvecsize{base} with  94M parameters and is already relatively large compared to other ASR models~\citep{Han2020ContextNetIC,Gulati2020ConformerCT}.
We study two strategies of changing the context network Transformer to reduce the model size and speed up the model, potentially at cost to performance: reducing model depth by using fewer Transformer layers  or reducing model width  by using smaller hidden state size in the Transformer.
When reducing the hidden size, we fix the head dimension to 64.
For example, a 12-head 768$d$ Transformer would be scaled down to a 4-head 256$d$ Transformer. 
The hidden size of the feed-forward network is always 4x as wide as the Transformer width. 
We also scale down the wave feature extractor to ensure that its width is not larger than the width of the Transformer.
For fairness of comparison, the depth counterpart uses the same wave feature extractor as well.

\shortautoref{fig:ablation_width_vs_depth} shows the performance-efficiency trade-offs of scaling down the depth or width of the model.
Scaling down the width achieves better performance-efficiency compared to scaling down the depth; a deep and narrow model is more favorable than a shallow and wide model.
These narrow models serve as the baselines for our following experiments.

\subsection{Temporal Resolution vs. Model Size}\label{subsec:resolution}

The wave feature extractor of W2V2 down-samples the raw audio to 50Hz frames with a stride size of 20ms, reducing the sequence length by a factor of 320 (\shortautoref{sec:w2v2}). 
However, even lower resolutions are common in prior end-to-end ASR approaches.
For example, several methods~\citep{Han2020ContextNetIC,Gulati2020ConformerCT} use log-mel filterbank features with a stride of 10ms (100Hz) and down-sample them to 40ms (25Hz) with two layers of strided convolutions. 
The result is halving the sequence length, and reducing the computation and memory footprint of the context network. 
Reducing the sequence length may allow increasing the model size with the same computation costs.

\begin{figure}[t]
    \centering
    \begin{minipage}[t]{.47\linewidth}
        \centering
        \includegraphics[width=\textwidth]{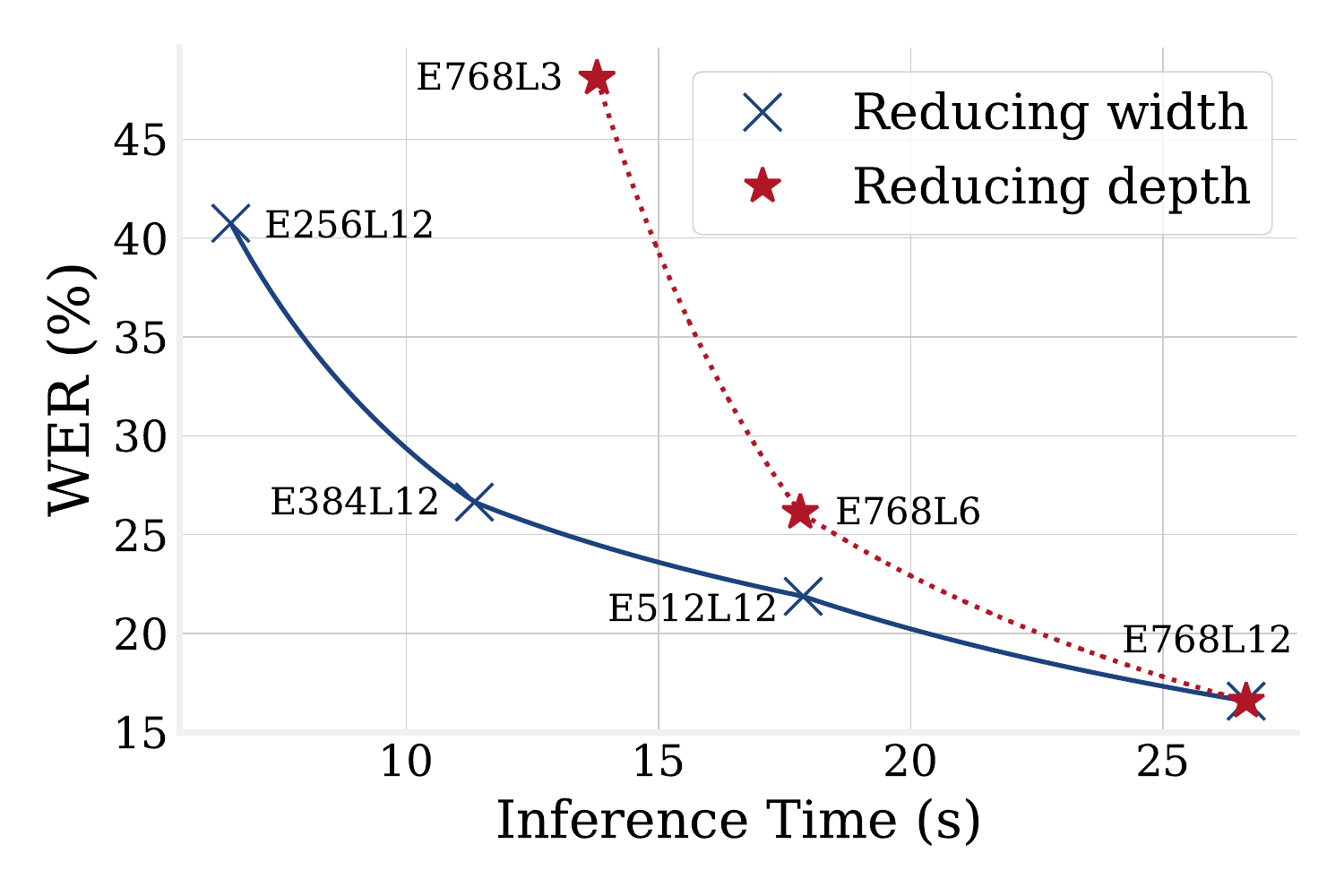}
        \caption{LibriSpeech dev-other WER versus inference time (with 100h labeled data).
        Reducing the width of the model (E768 $\rightarrow$ E512 $\rightarrow$ E384 $\rightarrow$ E256) achieves better performance-efficiency trade-off compared to reducing the depth (L12 $\rightarrow$ L6 $\rightarrow$ L3).
        }
        \label{fig:ablation_width_vs_depth}
    \end{minipage}
    \hspace{10pt}
    \begin{minipage}[t]{.47\linewidth}
        \centering
        \includegraphics[width=\linewidth]{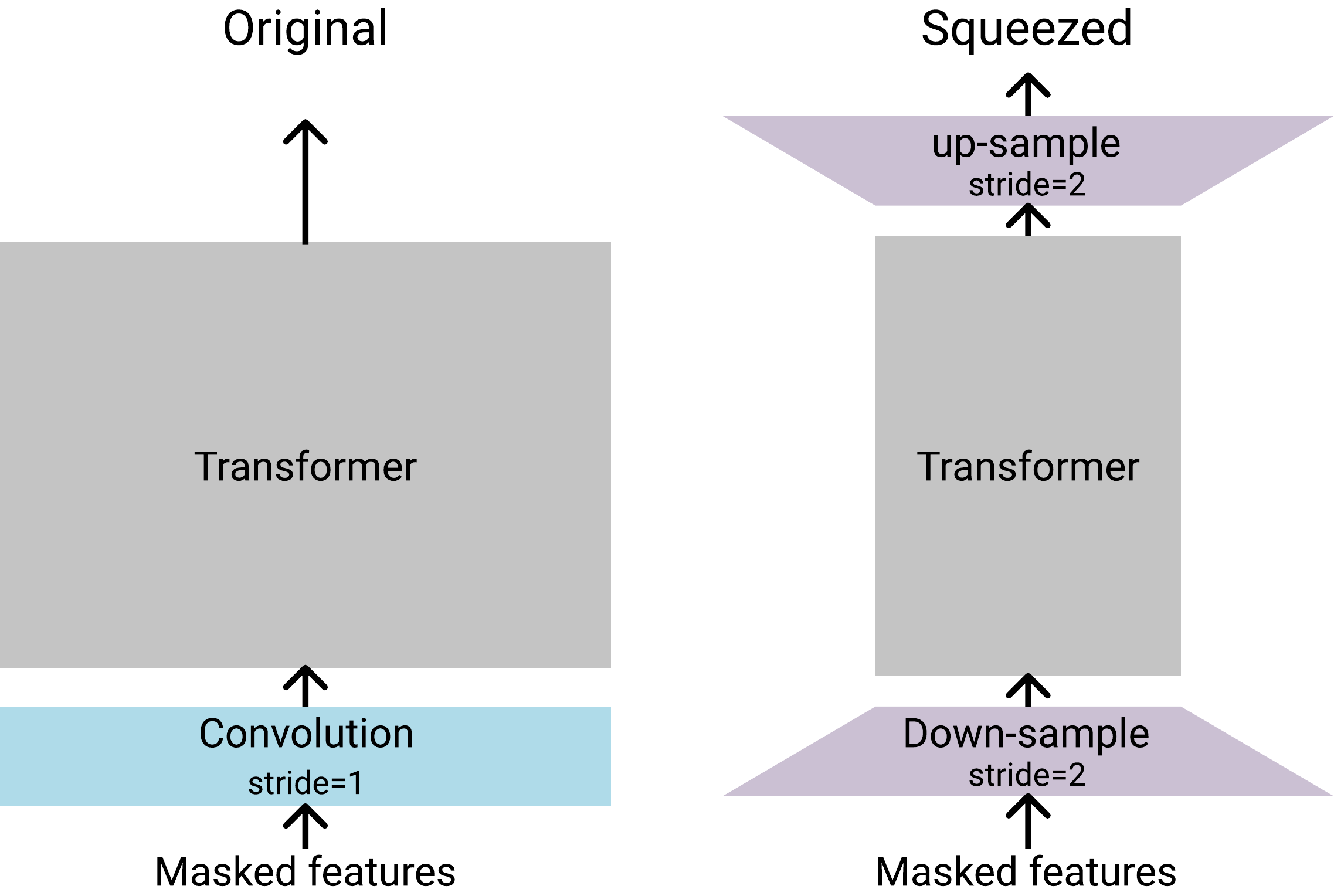} 
        \caption{Original vs. squeezed context network. The sequence length is halved by the down-sampling layer.}
        \label{fig:squeezed_cn}
    \end{minipage}
\end{figure}

\begin{table}[t]
\centering
\tablestyle{4pt}{1.1}
\begin{tabular}{lccccccc}
\toprule
& \multicolumn{3}{c}{Resolution} & & \multirow{2}{*}{Inference} & \multicolumn{2}{c}{dev-other} \\
\cmidrule(l{2pt}r{2pt}){2-4} \cmidrule(l{2pt}r{2pt}){7-8}
Model & Encode & Mask Pred. & CTC & \# Param. & Time (s) & WER & (+LM) \\
\midrule
\wavvec E256L12 (baseline) & 50Hz & 50Hz & 50Hz & 11.06M & 7.5\mypm{0.04} & 40.7 & 23.4 \\
\wavvec E384L12 25Hz & 25Hz & 25Hz & 25Hz & 23.76M & \textbf{6.7}\mypm{0.02} & 33.1 & 20.5 \\
\quad + FT upsample & 25Hz & 25Hz & 50Hz & 24.06M & \textbf{6.7}\mypm{0.02} & 31.5 & 18.5 \\
Sq-W2V E384L12 & 25Hz & 50Hz & 50Hz & 23.92M & 7.0\mypm{0.01} & \textbf{29.9} & \textbf{17.5} \\
\midrule
\wavvec E384L12 (WER lower bound) & 50Hz & 50Hz & 50Hz & 24.84M & \textcolor{red}{10.5}\mypm{0.03} & 28.3 & 16.8 \\
\bottomrule
\end{tabular}
\vspace{+3pt}
\caption{Comparing models with different resolutions at context encoding, mask prediction (for pre-training), and CTC (for ASR fine-tuning). The last row has the same WFE but a wider context network, which can be viewed as the lower bound of WER but has a much higher inference time (in \textcolor{red}{red}). The Squeezed wav2vec (Sq-W2V) closes the WER gap with similar inference time. We \textbf{bold} the best numbers except the lower bound.}\label{tab:ablation_resolution}
\tabbelowsqueeze
\end{table}

\shortautoref{tab:ablation_resolution} shows the performance-efficiency trade-off of models with different temporal resolutions at context encoding, mask prediction, and CTC decoding.
Reducing the temporal resolution while increasing the model size (first vs. second rows)  effectively reduces the WER while maintaining the inference time.
However, compared to a model with similar size but higher resolution (last row) there is a noticeable gap in WER.
Increasing the output resolution to 50Hz while keeping the encoding resolution the same (25Hz) (third row) reduces this gap. 
We added a transposed 1$d$ convolution layer\footnote{We use a Linear layer instead of a ConvTranspose1d in PyTorch~\citep{Paszke2019PyTorchAI} for efficiency.} to the output of the context network during fine-tuning, which allows each frame (25Hz) to generate two predictions (50Hz).

\paragraphsq{Squeezed Context Networks} 
To further close the gap, we propose  to  encode the features at a low resolution (e.g., 25Hz) while keeping contrastive learning at a high resolution (e.g., 50Hz).
We add a down-sampling layer and an up-sampling layer around the original context network.
Because there is already a convolution layer at the bottom of the \wavvec context network, we simply change its stride size from 1 to $s$ to avoid additional computation, where $s$ is the squeezing factor.\footnote{There is a shortcut connection in \wavvec --- it adds the inputs of the convolution to its outputs and passes it to the Transformer. We apply average pooling with kernel and stride sizes $s$ in this shortcut path which averages every two steps into one so that it can be added to the outputs of the strided convolution.}
The up-sample layer is a transposed 1$d$ convolution with kernel size $s$ and stride size $s$ ($s = 2$ in our experiments).
\shortautoref{fig:squeezed_cn} illustrates the context network squeezing.
The fourth row in \shortautoref{tab:ablation_resolution} shows that  using a squeezed context network  further reduces the WER with similar inference time.

\subsection{Wave Feature Extractors Design}

\wavvec has the same number of channels in all layers of its convolutional wave feature extractor (WFE-O; ``O'' stands for original).
\shortautoref{tab:wfe_flops} (left) shows FLOPs and inference time of a WFE-O with width 512.
The first few layers consume much of the computation time, while  the last three consume less than 10\% of the total computation. 
We hypothesize that the first few layers are unnecessarily large, and that the computation can be more evenly distributed across layers. 

\paragraphsq{Compact Wave Feature Extractors (WFE-C)}
We introduce a compact wave feature extractor (WFE-C) which doubles the number of channel when the sequence length is downsampled by 4 times. The progression of channel dimensionality is  ($c$, 2$c$, 2$c$, 4$c$, 4$c$, 8$c$, 8$c$) across its 7 conv layers where $c$ is a hyper-parameter. We keep the kernel sizes (7, 3, 3, 3, 3, 2, 2) and strides (5, 2, 2, 2, 2, 2, 2) of WFE-O.
\shortautoref{tab:wfe_flops} (right) shows the FLOPs and inference time of a WFE-C-c128-l0 (i.e., c = 128) feature extractor.
The inference time is distributed more evenly across layers.
\shortautoref{tab:ablation_wfe} presents the inference time and WER of a Squeezed wav2vec with different WFEs.
WFE-C-c128-l0 achieves a similar performance as WFE-O-c512 while being much faster.

\paragraphsq{WFEs Depth vs. Width}
We study scaling up WFE-C by adding a point-wise (kernel size 1) convolutional layer after each original convolutional layer except for the first layer, which creates a 13-layers convolutional network with kernel sizes (7, 3, 1, 3, 1, 3, 1, 3, 1, 2, 1, 2, 1) and strides (5, 2, 1, 2, 1, 2, 1, 2, 1, 2, 1, 2, 1). 
We refer this model as WFE-C-c128-l1, where ``l1'' denotes one additional intermediate layer between every two original layers.  
The last two rows of \shortautoref{tab:ablation_wfe} show the performance of increasing the width (WFE-C-c160-l0) and increasing the depth (WFE-C-c128-l1). 

\begin{table}[]
\centering
\tablestyle{4pt}{1.1}

\begin{tabular}{ccrrrcrrr}
\toprule
& \multicolumn{4}{c}{\textbf{WFE-O-c512}} & \multicolumn{4}{c}{\textbf{WFE-C-c128-l0}} \\
\cmidrule(l{2pt}r{2pt}){2-5} \cmidrule(l{2pt}r{2pt}){6-9}
& & & \multicolumn{2}{c}{Time (ms)} & &  & \multicolumn{2}{c}{Time (ms)} \\
\cmidrule(l{2pt}r{2pt}){4-5} \cmidrule(l{2pt}r{2pt}){8-9}
Layer & (c, k, s) & FLOPS (M) & CPU & GPU & (c, k, s) & FLOPS (M) & CPU & GPU \\
\midrule
0 & (512, 10, 5) & 197 & 4290 & 11.58 & (128, 10, 5) & 49 & 1071 & 4.18 \\
1 & (512, 3, 2) & 6356 & 9557 & 40.31 & (256, 3, 2) & 819 & 1913 & 6.70 \\
2 & (512, 3, 2) & 3178 & 5707 & 20.50 & (256, 3, 2) & 803 & 1486 & 6.56 \\
3 & (512, 3, 2) & 1588 & 2345 & 10.34 & (512, 3, 2) & 802 & 1376 & 5.03 \\
4 & (512, 3, 2) & 794 & 1181 & 5.21 & (512, 3, 2) & 794 & 1182 & 4.77 \\
5 & (512, 2, 2) & 266 & 464 & 1.69 & (1024, 2, 2) & 531 & 870 & 3.26 \\
6 & (512, 2, 2) & 133 & 215 & 0.88 & (1024, 2, 2) & 526 & 774 & 3.15 \\
\midrule
Total & & 12511 & 23759 & 90.51 & & 4325 & 8672 & 33.65 \\
\bottomrule
\end{tabular}
\vspace{+3pt}
    \caption{WFE-O-c512 vs. WFE-C-c128-l0 in terms of FLOPs and encoding time for a batch of twenty 10-second inputs. We show the number of channels (c), kernel size (k), and stride size (s) of each conv layer. WFE-C-c128-l0 allocates the FLOPs and processing time more evenly across layers.}\label{tab:wfe_flops}
\tabbelowsqueeze
\end{table}
\begin{table}[t]
    \centering
    \tablestyle{8pt}{1.1}
    \begin{tabular}{lccccc}
    \toprule
    WFE & \# Param. & PT Time (h) & Infer. Time (s) & WER & WER (+LM) \\
    \midrule
        WFE-O-c512 & 44.65M & 27.2 & 14.5\mypm{0.03} & 23.6 & 14.2 \\
    \midrule
    WFE-C-c128-l0 & 45.60M & \textbf{21.9} & \textbf{9.5}\mypm{0.01} & 23.8 & 14.0 \\
    WFE-C-c160-l0 & 48.33M & 24.3 & 11.0\mypm{0.03} & 23.5 & 14.2 \\
    WFE-C-c128-l1 & 48.35M & 24.4 & 11.2\mypm{0.00} & \textbf{22.8} & \textbf{13.6} \\
    \bottomrule
    \end{tabular}
    \vspace{+3pt}
    \caption{An E512L12 \wavvec with squeezed context network with different WFEs. WFE-C-c128-l0 performs similar to WFE-O-c512 while reducing overall model inference time by 37\%. With a similar inference time, depth increase (WFE-C-c128-l1) is better than width increase  (WFE-C-c160-l0).}
    \label{tab:ablation_wfe}
\tabbelowsqueeze
\end{table}

\subsection{Feature Extractor vs. Context Network}\label{subsec:wfe_vs_context}

We study where to allocate computation budgets: the feature extractor or the context network.
\shortautoref{tab:ablation_fix_infer_time} shows this study with a controlled inference time.
The third row is a model with a squeezed context network as we show in \shortautoref{subsec:resolution}.
The fourth row replaces WFE-O-c256 with WFE-C-c96-l1 and achieves better WER.
The fifth row reduces the size of the WFE and increases the size of its context network.
It outperforms the fourth row significantly in WER even as it shows a slightly lower inference time.
Moreover, we observe that \wavvec has a convolution layer with a particularly large kernel size of 128 at the bottom of the context network.
The sixth row shows a reduction of the size of this kernel to 31, which allows using a larger WFE-C.
It achieves similar WER as the fifth row while having lower pre-training and inference time.
We provide an additional ablation study of the kernel size in \shortautoref{app:sec:k127}.

\begin{table}[t]
\centering
\tablestyle{3pt}{1.1}
\begin{tabular}{clcccccc}
\toprule
&&  \multicolumn{2}{c}{\# Param. (M)} & \multicolumn{2}{c}{Time} & \multicolumn{2}{c}{Dev-other}\\
\cmidrule(l{2pt}r{2pt}){3-4} \cmidrule(l{2pt}r{2pt}){5-6} \cmidrule(l{2pt}r{2pt}){7-8}
\# & Model & All & WFE & PT (h) & Infer. (s) & WER & (+LM) \\
\midrule
1 & \wavvecsize{small} (E384L12 + WFE-O-c384) & 24.84 & 2.36 & 34.6 & 12.8\mypm{0.05} & 26.6 & 15.6  \\
2 & \wavvecsize{tiny} (E256L12 + WFE-O-c256) & 11.06 & 1.05 & 24.7 & 7.5\mypm{0.04} & 40.7 & 23.4 \\
\midrule
3 & Sq-W2V E384L12 + WFE-O-c256 & 23.92 & 1.05 & 23.4 & 7.0\mypm{0.01} & 29.9 & 17.5 \\
4 & Sq-W2V E384L12 + WFE-C-c96-l1 & 27.22 & 4.15 & 22.5 & 7.1\mypm{0.04} & 29.6 & 17.2 \\

5 & Sq-W2V E512L12 + WFE-C-c48-l1 & 41.69 & 1.04 & 21.7 & 7.1\mypm{0.01} & 24.4 & 14.7 \\
6 &  Sq-W2V E512L12 (conv k31) + WFE-C-c64-l1 & 40.73 & 1.84 & \textbf{19.1} & \textbf{6.7}\mypm{0.02} & 24.4 & 14.5 \\
7 & \; + MLP predictor heads (SEW-tiny) & 40.73 & 1.84 & 19.8 & \textbf{6.7}\mypm{0.02} & \textbf{23.7} & \textbf{14.4} \\
8 & \; - WFE-C-c64-l1 + FBFE-c160 & 42.15 & 3.26 & 19.2 & 7.5\mypm{0.02} & 25.0 & 15.1 \\
\bottomrule
\end{tabular}
\vspace{+3pt}
\caption{Ablation study with a controlled inference time. 
Rows 2 $\rightarrow$ 3: using a squeezed context network.
Rows 3 $\rightarrow$ 4: replace WFE-O with WFE-C.
Rows 4 $\rightarrow$ 5: Smaller WFE and larger Transformer.
Rows 5 $\rightarrow$ 6: Smaller conv kernel size in the context network and wider WFE.
Rows 6 $\rightarrow$ 7: using MLP predictor heads.
Rows 7 $\rightarrow$ 8: using 80D filter bank features (FBFE-c160, i.e. Filter Bank Feature Extractor with two 2D convolutional layers and 160 channels followed by a linear projection layer) instead of raw waveform (WFE-C).
}
\label{tab:ablation_fix_infer_time}
\tabbelowsqueeze
\end{table}

\subsection{MLP Predictor Heads}

\citet{Chen2020SimCLR} use MLP predictor heads instead of linear ones for unsupervised image representation learning, leading to better pre-trained features with little overhead during pre-training.
We replace the linear projection of \wavvec with a two-layer MLP with hidden size 4096, a ReLu activation in between, and BatchNorm~\citep{Ioffe2015BatchNA} after each linear layer.
Because the predictor heads are discarded following pre-training, there is no inference overhead.
The seventh row in \shortautoref{tab:ablation_fix_infer_time} shows the performance of using such an MLP predictor.
Compared to the sixth row, using such an MLP predictor leads to better performance without any additional inference time.
\shortautoref{app:sec:mlp} provides a more detailed ablation study on MLP predictor heads.

\subsection{Raw Waveform vs. Filter Bank Features Inputs}
\citet{Zhang2020PushingTL} propose a variant wav2vec 2.0 which removes the wave feature extractor and uses 80-dimensional log-mel filterbank (frame length 25ms and frame shift 10ms) and achieves superior performance with very large models. Their experiments apply other changes that can constitute confounding factors when interesting the results, including using the Conformer architecture~\citep{Gulati2020ConformerCT} and RNN-T~\citep{graves2012sequence} instead of CTC for ASR fine-tuning. We conduct additional experiments to evaluate the impact of using raw waveform inputs. The last row of \shortautoref{tab:ablation_fix_infer_time} shows the performance of our model using a FBFE (Filter Bank Feature Extractor) instead\footnote{Because \citet{Zhang2020PushingTL} do not provide implementation details, we borrow the publicly available implementation from ESPNet~\citep{watanabe2018espnet}, set the stride of the second convolution to 1 to match the encoding resolution, and reduce the number of channels to 160 to ensure the inference time is within the constraint.}. While using log-mel filter bank features can achieve reasonable performance, using raw waveform inputs with our WFE-C still achieves lower WER and faster inference time. 
\section{\finalmodel{} (\finalmodelfull)}\label{sec:final_model}

We combine our observations from \shortautoref{sec:method} to propose \textit{\finalmodel\ (\finalmodelfull)}, 
an efficient pre-trained model architecture. \finalmodel\ differs from  \wavvec in: 
(a) using a squezeed context network, (b) replacing WFE-O with WFE-C, (c) reallocating computing across different components, and (d) using MLP predictor heads with BatchNorm.
\shortautoref{tab:sew_config} shows the hyper-parameters of W2V2 and \finalmodel{} with different inference budgets, and \shortautoref{tab:ls_100h_100k} shows model performance. 

\begin{table}[]
\centering
\tablestyle{3pt}{1.1}
\begin{tabular}{lccccccccccccc}
    \toprule
          & \multicolumn{3}{c}{WFE} & \multicolumn{5}{c}{Context Network} & \multicolumn{2}{c}{Pred. Head} & \multicolumn{2}{c}{\# of Param. (M)}  &Infer. \\
          
    \cmidrule(l{2pt}r{2pt}){2-4} \cmidrule(l{2pt}r{2pt}){5-9} \cmidrule(l{2pt}r{2pt}){10-11} \cmidrule(l{2pt}r{2pt}){12-13}
    Model & Type & c & l & Conv-k & Sq. & E & L & D-Attn & Layer & BN & All & WFE & Time (s)\\
    \midrule
    \wavvecsize{tiny} & O & 256 & & 128 & & 256 & 12 & & 1 & & 11.1 & 1.1 & 7.5\mypm{0.04} \\
    \wavvecsize{small} & O & 384 & & 128 & & 384 & 12 & & 1 & & 24.8 & 2.4 & 12.8\mypm{0.05} \\
    \wavvecsize{mid} & O & 512 & & 128 & & 512 & 12 & & 1 & & 44.1 & 4.2 & 19.9\mypm{0.04} \\
    \wavvecsize{base} & O & 512 & & 128 & & 768 & 12 & & 1 & & 94.4 & 4.2 & 30.8\mypm{0.05} \\
    \wavvecsize{large} & O' & 512 & & 128 & & 1024 & 24 & & 1 & & 315.5 & 4.2 & 74.4\mypm{0.40} \\
    \midrule
    \finalmodelsize{tiny} & C & 64 & 1 & 31 & 2 & 512 & 12 & & 2 & \checkmark & 40.7 & 1.8 & 6.7\mypm{0.02} \\
    \finalmodelsize{small} & C & 64 & 1 & 31 & 2 & 768 & 12 & & 2 & \checkmark & 93.2 & 1.8 & 12.9\mypm{0.03} \\
    \finalmodelsize{mid} & C & 64 & 1 & 31 & 2 & 768 & 24 & & 2 & \checkmark & 178.2 & 1.8 & 21.0\mypm{0.04} \\
    \midrule
    \finalmodelDsize{tiny} & C & 64 & 1 & 31 & 2 & 384 & 12 & \checkmark & 2 & \checkmark & 25.0 & 1.8 & 8.5\mypm{0.03} \\
    \finalmodelDsize{small} & C & 64 & 1 & 31 & 2 & 512 & 12 & \checkmark & 2 & \checkmark & 42.6 & 1.8 & 10.9\mypm{0.03} \\
    \finalmodelDsize{mid} & C & 64 & 1 & 31 & 2 & 512 & 24 & \checkmark & 2 & \checkmark & 78.8 & 1.8 & 16.5\mypm{0.03} \\
    \finalmodelDsize{base} & C & 64 & 1 & 31 & 2 & 512 & 24 & \checkmark & 2 & \checkmark & 175.1 & 1.8 & 26.3\mypm{0.06} \\
    \finalmodelDsize{base+} & C & 96 & 1 & 31 & 2 & 512 & 24 & \checkmark & 2 & \checkmark & 177.0 & 4.1 & 27.8\mypm{0.05} \\
    \bottomrule
\end{tabular}
\vspace{+3pt}
\caption{Model hyper-parameters, categorized by inference time. We focus on performance-efficiency trade-off, and therefore we do not control for model size within each time category. O': \wavvecsize{large} removes the InstanceNorm layer after the first convolution and adds LayerNorm after each convolution. }
\label{tab:sew_config}
\tabbelowsqueeze
\end{table}

\paragraphsq{Scaling Up}
We adopt a simple scaling up recipe, leaving finding more optimal scaled-up configurations, an open research problem~\citep{Tan2019EfficientNetRM,Dollr2021FastAA}, for future work. 
We take the row 7 in \autoref{tab:ablation_fix_infer_time} as \finalmodelsize{tiny}, which has similar inference times to \wavvec with width 256.
We increase the width by 1.5 to create \finalmodelsize{small}, which has the same Transformer size as \wavvecsize{base}.
Based on our observation that deep models are favorable (\shortautoref{subsec:depth_vs_width}), we create \finalmodelsize{mid} by making the model twice as deep.

\paragraphsq{\finalmodelD (\finalmodel\ with Disentangled Attention)}

Disetangled attention~\citep{He2020DeBERTaDB} is a variant of self-attention with relative position representations~\citep{Shaw2018SelfAttentionWR}, which outperforms  Transformer's multi-head attention~\citep{Vaswani2017AttentionIA} on various NLP tasks.
Unlike Transformer which adds absolute positional embeddings to the content embeddings at the beginning, disentangled attention keeps the positional embeddings and content embeddings separate and has 3 components in its attention weight computation: (a) content-to-content, (b) content-to-position, and (c) position-to-content attentions (\shortautoref{app:sec:disent_attn}). 
The disentangled computation requires more matrix multiplication operations compared to conventional self-attention, and is slower with a similar number of parameters.
To retain similar computation costs, we reduce the Transformer width to reduce its parameters by half. 
\finalmodelD benefits from the advanced attention mechanism, while overall it displays faster inference time. 
In \shortautoref{sec:exp}, we show that \finalmodelD outperforms a 2x larger \finalmodel~counterpart.
To have a model with comparable inference time as \wavvecsize{base}, we further increase the width of the context network of \finalmodelD by 1.5 to create \finalmodelDsize{base}. Because it is still slightly faster than \wavvecsize{base}, we further scale up the width of the WFE by 1.5 leading to a \finalmodelDsize{base+}.

\section{Further Experiments}\label{sec:exp}

We compare \finalmodel and \finalmodel-D to W2V2 using a variety of fine-tuning setups.

\begin{table}[t]
\centering
\tablestyle{4pt}{1.1}
\begin{tabular}{lrcccccc}
\toprule
& & \multicolumn{2}{c}{Time} & \multicolumn{4}{c}{WER (No LM / 4-gram LM beam=50)} \\
\cmidrule(l{2pt}r{2pt}){3-4}
\cmidrule(l{2pt}r{2pt}){5-8}
Model & \# Param. & PT (h) & Infer. (s) & dev-clean & dev-other & test-clean & test-other \\
\midrule

\wavvecsize{tiny} & 11.1 & 24.7 & 7.5\mypm{0.04} & 22.0 / 7.7 & 40.7 / 23.4 & 22.8 / 8.3 & 42.1 / 25.6 \\
\finalmodelsize{tiny}  & 40.7 & \textbf{19.8} &  \textbf{6.7\mypm{0.02}} & 10.6 / 4.6 & 23.7 / 14.4 & 10.6 / 5.1 & 23.7 / 14.5 \\
\finalmodelDsize{tiny}  & 24.1 & 23.8 & 7.5\mypm{0.02} & \textbf{10.1} / \textbf{4.4} & \textbf{22.3} / \textbf{13.4} & \textbf{10.4} / \textbf{4.9} & \textbf{22.8} / \textbf{13.9} \\
\midrule
\wavvecsize{small} & 24.8 & 34.6 & 12.8\mypm{0.05} & 12.4 / 5.0 & 26.6 / 15.6 & 12.8 / 5.7 & 27.2 / 16.0 \\
\finalmodelsize{small} & 89.6 & \textbf{32.3} & 11.0\mypm{0.02} & 7.6 / \textbf{3.6} & \textbf{17.5} / \textbf{10.9} & \textbf{7.8} / \textbf{4.2} & \textbf{18.0} / 11.5 \\
\finalmodelDsize{small} & 41.0 & 38.1 & \textbf{9.6\mypm{0.04}} & \textbf{7.5} / \textbf{3.6} & 17.9 / 11.1 & \textbf{7.8} / \textbf{4.2} & 18.2 / \textbf{11.4} \\
\midrule
\wavvecsize{mid} & 44.1 & \textbf{40.3} & 19.9\mypm{0.04} & 9.3 / 4.1 & 21.9 / 13.2 & 9.6 / 4.8 & 22.2 / 13.5 \\
\finalmodelsize{mid} & 174.7 & 41.7 & 19.1\mypm{0.03} & 6.5 / 3.4 & 14.7 / 9.5 & 6.7 / 3.9 & 14.9 / 10.0 \\
\finalmodelDsize{mid} & 78.8 & 51.7 & \textbf{16.5\mypm{0.03}} & \textbf{6.3} / \textbf{3.2} & \textbf{14.0} / \textbf{9.3} & \textbf{6.4} / \textbf{3.8} & \textbf{14.2} / \textbf{9.5} \\
\midrule
\wavvecsize{base} & 94.4 & \textbf{55.2} & 30.8\mypm{0.05} & 6.9 / 3.4 & 16.6 / 10.4 & 7.1 / 4.0 & 16.4 / 10.4 \\

\finalmodelDsize{base} & 175.1 & 59.1 & \textbf{26.3\mypm{0.06}} & 5.8 / 3.2 & 12.6 / 8.6 & 5.8 / 3.6 & 13.2 / 9.3 \\
\finalmodelDsize{base+} & 177.0 & 68.4 & 27.8\mypm{0.05} & \textbf{5.3} / \textbf{3.0} & \textbf{12.4} / \textbf{8.7} & \textbf{5.3} / \textbf{3.5} & \textbf{12.6} / \textbf{9.0} \\
\bottomrule

\end{tabular}
\vspace{+3pt}
\caption{Libri-Speech 100h-960h semi-supervised setup pretrained for 100K updates.}
\label{tab:ls_100h_100k}
\tabbelowsqueeze
\end{table}

\paragraphsq{\finalmodel{} vs. \finalmodelD vs. \wavvec  on LibriSpeech 100h-960h}
We pre-train \wavvec, \finalmodel, and \finalmodelD on 960h LibriSpeech audios for 100K updates and fine-tune them on 100h labelled data. We follow the setup of \shortautoref{subsec:exp_setup}. 
\shortautoref{tab:ls_100h_100k} shows pre-training times, inference times, and WERs with and without an LM.
Without an LM, compared with the \wavvecsize{tiny}, \finalmodelsize{tiny} reduces the WER by 53.5\% (22.8\% to 10.6\%) and 43.7\% (41.1\% to 23.7\%) on test-clean and test-other, while being faster.
With an LM, WER improves by 38.6\% and 43.4\% on test-clean and test-other.
Compared with the \wavvecsize{mid}, \finalmodelsize{mid} reduces WER by 30.2\% (9.6\% to 6.7\%) and 32.9\% (22.2\% to 14.9\%) with similar inference times. 
\finalmodel does incur slight increase in training time compared to \wavvec with similar inference times.
However, \finalmodel has lower WER even compared to a slower \wavvec which takes longer to train (e.g., \finalmodelsize{small} vs. \wavvecsize{mid} or \finalmodelsize{mid} vs. \wavvecsize{base}).

\finalmodelD has lower WER compared to \finalmodel{} even with smaller width and half of the parameters. With large models, \finalmodelD is also more efficient. 
However, \finalmodelDsize{tiny} is slower than \finalmodelsize{tiny}, due to the implementation difference.\footnote{We use the official PyTorch~\citep{Paszke2019PyTorchAI} implementation of disentangled attention (\href{https://github.com/microsoft/DeBERTa}{https://github.com/microsoft/DeBERTa}) for \finalmodelD, which uses BTC tensor format instead of a more efficient TBC tensor format used in fairseq~\citep{Ott2019fairseqAF}. Moreover, fairseq uses a dedicated CUDA implementation of self-attention which is more efficient.}

\begin{table}[t]
\centering
\tablestyle{4pt}{1.1}
\begin{tabular}{lccccccc}
\toprule
& Inference & \multicolumn{2}{c}{10m} & \multicolumn{2}{c}{1h} & \multicolumn{2}{c}{10h} \\
\cmidrule(l{2pt}r{2pt}){3-4}
\cmidrule(l{2pt}r{2pt}){5-6}
\cmidrule(l{2pt}r{2pt}){7-8}
Model & Time (s) & test-clean & test-other & test-clean & test-other & test-clean & test-other \\
\midrule

\wavvecsize{base} & 30.8\mypm{0.05} & 47.8\mypm{1.16} & 54.0\mypm{1.18} & 33.3\mypm{2.73} & 40.0\mypm{2.34} & 13.3\mypm{0.27} & 22.6\mypm{0.15} \\
 + LM & & \textbf{19.7}\mypm{0.26} & \textbf{28.2}\mypm{0.23} & 9.9\mypm{0.48} & 18.4\mypm{0.65} & 6.4\mypm{0.08} & 13.8\mypm{0.13} \\
\midrule
\finalmodelDsize{mid} & \textbf{16.5}\mypm{0.03} & 54.2\mypm{1.36} & 59.5\mypm{1.09} & 30.4\mypm{0.08} & 36.8\mypm{0.18} & 12.3\mypm{0.06} & 20.7\mypm{0.07} \\
+ LM & & 25.5\mypm{1.06} & 33.4\mypm{1.20} & 9.6\mypm{0.08} & 17.5\mypm{0.11} & 5.9\mypm{0.04} & 13.0\mypm{0.07} \\
\midrule
\finalmodelDsize{base+} & 27.8\mypm{0.05} & \textbf{43.4}\mypm{0.54} & \textbf{49.3}\mypm{0.42} & \textbf{21.5}\mypm{0.22} & \textbf{28.9}\mypm{0.11} & \textbf{11.0}\mypm{0.35} & \textbf{18.1}\mypm{0.06} \\
+ LM & & 22.6\mypm{0.40} & 30.6\mypm{0.25} & \textbf{9.5}\mypm{0.37} & \textbf{17.4}\mypm{0.38} & \textbf{5.2}\mypm{0.06} & \textbf{11.7}\mypm{0.24} \\
\bottomrule
\end{tabular}
\vspace{+3pt}
\caption{LibriSpeech results with 10m, 1h, or 10h supervised data. All the models are pre-trained for 100K updates on LibriSpeech 960h. We report inference times on dev-clean and WERs without LM and with 4-gram LM (beam size 50). The mean and standard deviations are calculated from three random runs. Similar to BERT-large~\citep{phang2018sentence}, we observe that our \finalmodelDsize{base+} is unstable during fine-tuning. We run five random runs and stop two degenerate runs after 3K updates.}
\label{tab:ls_1m_1h_10h}
\tabbelowsqueeze
\end{table}
\begin{table}[t]
\centering
\tablestyle{4pt}{1.1}
\begin{tabular}{lrcccccc}
\toprule
& & \multicolumn{2}{c}{Time} & \multicolumn{4}{c}{WER (No LM / 4-gram LM beam=50)} \\

\cmidrule(l{2pt}r{2pt}){3-4}
\cmidrule(l{2pt}r{2pt}){5-8}
Model & \# Param. & PT (GPU-day) & Infer. (s) & dev-clean & dev-other & test-clean & test-other \\
\midrule
W2V2-base & 94.4 & 102.4(73.5) & 30.8\mypm{0.05} & 6.1 / 2.9 & 13.6 / 8.5 & 6.1 / 3.5 & 13.3 / 8.7 \\
W2V2-large & 315.5 & 294.4 & 74.4\mypm{0.40}  & 4.6 / \textbf{2.4} & 9.3 / \textbf{6.1} & 4.7 / \textbf{2.9} & \textbf{9.1} / \textbf{6.4} \\
\midrule
SEW-D-mid & 78.8 & \textbf{68.9} & \textbf{16.5\mypm{0.03}} & 4.8 / 2.7 & 11.1 / 7.6 & 4.9 / 3.3 & 11.5 / 8.2 \\
SEW-D-base+ & 177.0 & 91.2 & 27.8\mypm{0.05} & \textbf{4.1} / 2.5 & \textbf{9.2} / 6.6 & \textbf{4.4} / 3.1 & 9.2 / 7.0 \\
\bottomrule
\end{tabular}
\vspace{+3pt}
\caption{LibriSpeech 100h-960h semi-supervised setup pretrained for 400K updates. We use the public W2V2s~\citep{baevski2020wav2vec2} as the baselines. Notably, W2V2-base is reported as taking 102.4 GPU-days to pre-train on 64 GPUs, but from our estimation it only takes 73.5 GPU-days on 8 GPUs. Unlike \citet{baevski2020wav2vec2}, we neither tune the decoding hyper-parameters nor use a huge beam size 1500.}
\label{tab:ls_100h_400k}
\tabbelowsqueeze
\end{table}

\paragraphsq{Less Supervision}
To further test the performance of  \finalmodelD, we experiment with only 10min, 1h, and 10h of supervised data (see \shortautoref{app:sec:exp_setup_details}).
\shortautoref{tab:ls_1m_1h_10h} shows the WER of \wavvec and \finalmodelD.
\finalmodelDsize{mid} outperforms \wavvecsize{base} in the 1h and 10h scenarios while being more efficient. \finalmodelDsize{mid} is worse than \wavvecsize{base} in the extreme 10m scenario; however, we did not tune the fine-tuning hyper-parameters and use the ones tuned for \wavvecsize{base}.
\finalmodelDsize{base+} achieves significantly better performance than \wavvecsize{base} in most of the setups except for using 10 minutes supervision and decoded with LM. Potentially due to a large model size, we observe that \finalmodelDsize{base+} is unstable to fine-tune; therefore, instead of using \wavvec's tuned hyperparameters, we reduce the learning rate by 5 times to $10^{-5}$, set the dropout rates as the pre-training (\wavvec uses different sets of dropouts during pre-training and fine-tuning), and do not freeze the context network at the beginning of the fine-tuning. These adjustments stabilizes the fine-tuning of the model.

\paragraphsq{Comparison to Published Results}
We continue training our best \finalmodelDsize{mid} model to 400K updates and compare it with the official \wavvecsize{base}\footnote{\href{https://dl.fbaipublicfiles.com/fairseq/wav2vec/wav2vec_small_100h.pt}{https://dl.fbaipublicfiles.com/fairseq/wav2vec/wav2vec\_small\_100h.pt}} and \wavvecsize{large}\footnote{\href{https://dl.fbaipublicfiles.com/fairseq/wav2vec/wav2vec_small_100h.pt}{https://dl.fbaipublicfiles.com/fairseq/wav2vec/wav2vec\_big\_100h.pt}} checkpoints~\citep{baevski2020wav2vec2}.
\shortautoref{tab:ls_100h_400k} shows  inference times and WERs with or without an LM.
Compared to \wavvecsize{base}, \finalmodelDsize{mid} reduces inference time by 46.4\% (a 1.9$\times$ speed-up) and WER by 19.7\% and 13.5\% on both test sets without an LM; \finalmodelDsize{base+} reduces inference time by 9.7\% and WER by 27.9\% and 30.8\%. Compared to \wavvecsize{large}, \finalmodelDsize{base+} achieves 2.7$\times$ and 3.2$\times$ speed-ups for inference and pre-training with comparable WER and half of the number of parameters.

\paragraphsq{Transferring to Out-of-domain Data}

\begin{table}[t]
\centering
\tablestyle{3pt}{1.1}
\scriptsize
\begin{tabular}{lccccccccc}
\toprule
& \multicolumn{3}{c}{TED-LIUM 3 (10h)} & \multicolumn{3}{c}{VoxPopuli (10h)} & \multicolumn{3}{c}{Fisher+Switchboard (10h)} \\
\cmidrule(l{2pt}r{2pt}){2-4}
\cmidrule(l{2pt}r{2pt}){5-7}
\cmidrule(l{2pt}r{2pt}){8-10}
Model & Time (s) & dev (\%) & test (\%) & Time (s) & dev (\%) & test (\%) & Time (s) & dev (\%) & test (\%) \\
\midrule
\wavvecsize{base} & 10.2\mypm{0.03} & 15.7\mypm{0.08} & 14.8\mypm{0.13} & 30.8\mypm{0.22} & 17.5\mypm{0.10} & 18.1\mypm{0.03} & 33.8\mypm{0.12} & 23.2\mypm{0.06} & 23.2\mypm{0.10} \\
+ LM & & 9.2\mypm{0.09} & 9.3\mypm{0.09} & & 11.2\mypm{0.07} & 11.6\mypm{0.01} & & 16.5\mypm{0.06} & 16.7\mypm{0.00} \\
\midrule
\finalmodelDsize{mid} & 6.9\mypm{0.02} & 14.4\mypm{0.11} & 13.9\mypm{0.04} & 20.3\mypm{0.08} & 17.5\mypm{0.10} & 18.1\mypm{0.03} & 21.6\mypm{0.07} & 24.1\mypm{0.67} & 24.3\mypm{0.49} \\
+ LM & & 9.0\mypm{0.24} & 9.2\mypm{0.11} & & 10.9\mypm{0.22} & 11.2\mypm{0.26} & & 18.0\mypm{0.44} & 18.3\mypm{0.53} \\
\midrule
\finalmodelDsize{base+} & 11.2\mypm{0.01} & \textbf{12.8}\mypm{0.28} & \textbf{12.2}\mypm{0.14} & 33.8\mypm{0.04} & \textbf{14.8}\mypm{0.27} & \textbf{15.5}\mypm{0.35} & 36.8\mypm{0.03} & \textbf{20.5}\mypm{0.10} & \textbf{20.5}\mypm{0.06} \\
+ LM & & \textbf{8.3}\mypm{0.39} & \textbf{8.7}\mypm{0.22} & & \textbf{10.3}\mypm{0.13} & \textbf{10.9}\mypm{0.39} & & \textbf{15.6}\mypm{0.21} & \textbf{15.7}\mypm{0.06} \\
\bottomrule
\end{tabular}
\vspace{+3pt}
\caption{Inference time and WER of LibriSpeech pre-trained models (400K updates) transferred to TED-LIUM 3, VoxPopuli, and Fisher+Switchboard datasets with only 10h labels. \finalmodel-D mid outperforms W2V2 base on all settings while being at least 30\% faster. We report the inference time on the dev sets. The mean and standard deviations are computed over three random runs.}
\label{tab:transfer_results_400k}
\tabbelowsqueeze
\end{table}

We evaluate \wavvec and \finalmodelD pre-trained models on three additional ASR datasets: TED-LIUM 3 (CC BY-NC-ND 3.0)~\citep{Hernandez2018TEDLIUM3T}, VoxPopuli (CC0, CC BY-NC 4.0)~\citep{Wang2021VoxPopuliAL}, and Fisher+Switchboard (LDC200\{4,5\}S13, LDC200\{4,5\}T19, LDC97S62)~\citep{switchboard,fisher-a,fisher-b} with a similar setup to \citet{Hsu2021RobustW2} (see \shortautoref{app:sec:exp_setup_details}).
We use only 10h of supervised audio to stress test low resource domain transfer.
\shortautoref{tab:transfer_results_400k} shows the inference times and WERs.
\finalmodelDsize{mid} consistently reduces inference times by about 30\% while providing lower WERs on TED-LIUM 3, similar WERs on VoxPopuli, and slightly higher WERs on Fisher+Switchboard. \finalmodelDsize{base+} consistently outperforms \wavvecsize{base} by a large margin while being only 10\% slower.

\section{Conclusion}\label{sec:conclusion}

Our study is a detailed analysis of the architecture of \wavvec, an influential pre-training method for spoken language tasks. 
Through careful consideration of both compute time and model expressivity we achieve better ASR performance with faster inference times. 
Aggregating our observations, we propose \finalmodel, a family of pre-trained models, with significantly better performance-efficiency trade-off than the existing \wavvec architecture. 
\finalmodel models can function as direct replacement of \wavvec models, including in recent work~\citep{hsu2020hubert,Hsu2021RobustW2,Xu2020SelftrainingAP}.

In general, our approach outlines a recipe and set of considerations to apply when studying complex network architectures with the goal of finding better balance of performance and efficiency. 
While model performance is commonly prioritized in research, the economics of inference times are often as critical for model deployment in the real world. 
This study will inform practitioners optimizing complex models for deployment beyond this specific instance of \wavvec.

{\small
\bibliography{ref}
\bibliographystyle{plainnat}
}

\clearpage

\appendix
\section{Limitation}\label{app:sec:limitation}

We focus on model inference time, without considering LM computation times.
As discussed in \shortautoref{app:sec:ctc_decoding}, we perform beam-search decoding with LM on CPU using \wavvec's implementation. This results in significant slowdown of tiny models.
How to speed it up is an important direction for future work.
Additionally, different hardware devices require different optimized models.
All our ablation studies are done on GPUs. 
Our observation may change on other types of hardware, such as embedded systems or CPUs. 
As we discuss in \shortautoref{app:sec:social_impact}, there remain a need to study ASR across more diverse types of data (e.g., across languages, domains, ethnic groups, etc.). Similar to existing work, which we compare with, different data may lead to different performance observations. However, we do not expect significant changes in computation speedups, the main focus of our work. 

\section{Experimental Setup Details}
\label{app:sec:exp_setup_details}

\subsection{LibriSpeech}
LibriSpeech (CC BY 4.0)~\citep{Panayotov2015LibrispeechAA} is a corpus of 16kHz read-English speech. The data are derived from read audiobooks from the LibriVox project.
LibriSpeech includes a 960h training set (including three subsets: train-clean-100, train-clean-360, and train-other-500), two development sets, dev-clean (5.4h) and dev-other (5.3h), and two test sets, test-clean (5.4h) and test-other (5.1h). 
The data is designed to provide more challenging evaluation with the dev-other and test-other splits. 
We use train-clean-100 as the 100h supervised data.
For 10min, 1h, and 10h subset of labelled data, we use splits provided by Libri-Light~\citep{librilight}.\footnote{\url{https://dl.fbaipublicfiles.com/librilight/data/librispeech_finetuning.tgz}}

\paragraphsq{Pre-training}
We use \wavvec's official codebase as provided in fairseq~\citep{Ott2019fairseqAF} for all experiments.
Following the provided configuration\footnote{\url{https://github.com/pytorch/fairseq/blob/master/examples/wav2vec/config/pretraining/wav2vec2_base_librispeech.yaml}}, we use the Adam~\citep{Kingma2015AdamAM} optimizer with learning rate 0.0005, betas (0.9, 0.98), weight decay 0.01, and 32K linear warm-up steps~\citep{Goyal2017AccurateLM}. We apply Layerdrop~\citep{Huang2016DeepNW,Fan2020ReducingTD} of 0.05.
We use layerdrop rate of 0.1 and 0.2 for \finalmodel and \finalmodelD. Otherwise, the models diverge. This follows the configuration of \wavvecsize{large} which also has 24 Transformer layers. 
The learning rate is decayed linearly to 0 from 32K steps to 400K steps.
Time masking probability is set to 0.065 with mask length 10.
Audio examples are batched by length to ensure that the total batch (across 64 GPUs) contains at most $5,600 = 64 \times 1,400,000 / 16,000$ seconds of audios.
We use 8 gradient accumulation steps to simulate 64-GPU training with 8 GPUs.
Each GPU processes at most $87.5 = 1,400,000 / 16,000$ seconds of audio in each forward-backward pass.
For tiny models, the memory usage is lower, so we double this number to 175 seconds and half the gradient accumulation steps to 4, which makes GPUs less underutilized and shortens the pre-training time.
This modification does not change the maximum size of the total batch size.
We use half-precision (FP16) training.

\paragraphsq{Fine-tuning on 10m or 1h Supervised Labels} 
We use the Adam optimizer with learning rate $3 \times 10^{-5}$, betas (0.9, 0.98) and tri-stage learning rate scheduler~\citep{Zhang2020TransformerTA} (10\% warm-up, 40\% constant, 50\% exponential decay to 5\% of the peak learning rate) for 13K updates .
The models are fine-tuned for 13K updates. In the first 10K updates, the context network is frozen and only the additional linear layer is trained. The WFE is always frozen.
The audios are batched by length to ensure that the total batch (across 8 GPUs) contains at most $1,600 = 8 \times 3,200,000 / 16,000 $ seconds of audios.
We set gradient accumulation steps to 8 to simulate 8-GPU fine-tuning using a single GPU.

\paragraphsq{Fine-tuning on 10h Supervised Labels}
The models are fine-tuned for 20K updates. In the first 10K updates, the context network is frozen and only the additional linear layer is trained. The WFE is always frozen.
The rest of the settings are the same as the 10-minute scenario.

\paragraphsq{Fine-tuning on 100h Supervised Labels}
The models are fine-tuned for 80K updates. The context network is fine-tuned from the beginning (i.e., not frozen).
The WFE is frozen all the time.
The rest of the settings are the same as the 10-minute scenario.
To speed up the experiment, we set gradient accumulation steps to 2 to simulate fine-tuning with 4 GPUs.

\paragraphsq{Inference}
Without an LM, we decode the models with a CTC greedy decoding (a.k.a. best path decoding) algorithm, which takes the most likely token at each timestep, collapses the duplicated tokens, and removes all blank tokens.
When using an LM, we use the wav2letter lexicon decoder~\citep{Collobert2016Wav2LetterAE} which uses beam search. We set the beam size to 50, LM weight to 2, and word score to -1.
We use the official lexicon and a 4-gram LM.\footnote{\url{https://www.openslr.org/11/}}
During decoding the audio examples are sorted by length and batched with at most $250 = 4,000,000 / 16,000$ seconds in a batch.
All these hyper-parameters are the default values in \wavvec's official inference code.
The inference time is estimated on an AWS p3.2xlarge instance with 1 NVIDIA V100-SXM2-16GB GPU and 8 Intel(R) Xeon(R) CPU E5-2686 v4 @ 2.30GHz.
We run 5 trials and report the mean and standard deviation of the inference time.

\subsection{TED-LIUM 3}
TED-LIUM 3 (CC BY-NC-ND 3.0)~\citep{Hernandez2018TEDLIUM3T} is an English speech recognition corpus extracted from the public TED talks.
It includes a 452h training set, a 1.6h development set, and a 2.6h test set. We follow the Kaldi~\citep{kaldi} data preparation recipe.\footnote{\url{https://github.com/kaldi-asr/kaldi/tree/master/egs/tedlium/s5_r3}}
We randomly sample the 10h labelled data from the training set.

\paragraphsq{Fine-tuning on 10h Supervised Labels}
We following the same fine-tuning hyperparameters as the LibriSpeech 10h setup.

\paragraphsq{Inference}
We follow \citet{Hsu2021RobustW2} to create a 5-gram LM. Otherwise, we use the same inference setup for LibriSpeech.

\subsection{VoxPopuli}
VoxPopuli (CC0, CC BY-NC 4.0)~\citep{Wang2021VoxPopuliAL} is a large-scale multilingual speech corpus collected from 2009--2020 European Parliament event recordings. We use the transcribed English corpus which consists of a 522h training set, a 5h development set, and a 5h test set. We randomly sample the 10h labelled data from the training set.

\paragraphsq{Fine-tuning on 10h Supervised Labels} We follow the same fine-tuning hyperparameters as the LibriSpeech 10h setup.

\paragraphsq{Inference}
We use the official lexicon and 5-gram LM.\footnote{\url{https://github.com/facebookresearch/voxpopuli}}
Otherwise, we use the same inference setup for LibriSpeech.

\subsection{Fisher+Switchboard}
Fisher (LDC200\{4,5\}S13, LDC200\{4,5\}T19)~\citep{fisher-a,fisher-b} and Switchboard (LDC97S62)~\citep{switchboard} are conversational telephone speech corpora recorded in 8kHz. We combine them to create a 2250h training set. RT-03S (LDC2007S10, 6.3h)~\citep{rt-03s} and Hub5 Eval2000 (LDC2002S09, 3.6h)~\citep{eval2000} are used as  a development set and a test set. We preprocess the data according to \citet{Hsu2021RobustW2} including re-sampling 8kHz data to 16kHz. We use the Kaldi~\citep{kaldi}  data preparation and evaluation recipe.\footnote{\url{https://github.com/kaldi-asr/kaldi/tree/master/egs/fisher_swbd/s5}}
We randomly sample the 10h labelled data from the training set.

\paragraphsq{Fine-tuning on 10h Supervised Labels} 
We use the same fine-tuning hyperparameters as the LibriSpeech 10h setup.

\paragraphsq{Inference}
We follow \citet{Hsu2021RobustW2} to create a 4-gram LM using all the texts in the training set.
Otherwise, we use the same inference setup for LibriSpeech.
\section{Ablation Study on MLP Predictor Heads} \label{app:sec:mlp}

\shortautoref{tab:ablation_mlp_heads} shows LibriSpeech 100h performance with various prediction heads. 
Similar to vision models, batch normalization in the prediction heads improves \wavvec performance.
Unlike vision models where the prediction head is applied to merely a single pooled vector for each example, in \wavvec, the prediction heads are applied to all timesteps, which leads to higher pre-training overheads, but still  no overhead during fine-tuning or inference where the prediction heads are dropped.

\begin{table}[t]
    \centering
    \tablestyle{8pt}{1.1}
    \begin{tabular}{lcccc}
        \toprule
        Predictor Heads & \# Param. (PT) & PT time (hr) & WER & WER (+ LM) \\
        \midrule
        Linear (baseline) & 44.7M & \textbf{40.3} & 21.9 & 13.2 \\
        \midrule
        2-layer MLP & 49.8M & 42.0 & 20.4 & 12.4 \\
        2-layer MLP + BN & 49.8M & 42.9 & \textbf{20.0} & \textbf{12.2} \\
        3-layer MLP + BN & 83.4M & 45.3 & \textbf{20.0} & 12.3 \\
        \bottomrule
    \end{tabular}
    \vspace{3pt}
    \caption{LibriSpeech test-other results using different MLP predictor heads. For simplicity, we use a \wavvec E512L12 (pre-trained on 960h, fine-tuned on 100h) as  baseline. All models have the same inference time and number of parameters (44.1M) because the predictor heads are discarded during fine-tuning on the downstream tasks. Using a 2-layer MLP with BatchNorm performs the best while adding only $7\%$ to the pre-training (PT) time.}
    \label{tab:ablation_mlp_heads}
    \tabbelowsqueeze
\end{table}
\section{CTC Decoding} \label{app:sec:ctc_decoding}

\paragraphsq{Best-path (greedy) Decoding vs. Viterbi Decoding}

Best path decoding finds the most likely decoding path. Viterbi decoding~\citep{Viterbi1967ErrorBF} finds the most likely collapsed token sequence by summing up all possible paths.
Since the outputs of a CTC model at different timesteps are independent, the best path can be generated by taking the most likely token at each timestep.
This makes decoding parallelizable and efficient on GPUs.
The Viterbi decoding algorithm is a dynamic programming algorithm, which processes the model outputs sequentially and is implemented on CPUs in \wavvec's implementation.
\shortautoref{tab:greedy_vs_viterbi} shows the WER and inference time using best-path (greedy) and Viterbi decoding methods using four official \wavvec ASR models.
Two method achieve exactly the same WER while the best path decoding is faster.
Although the difference is only about one second, it is significant when using tiny models.
It implies that all models are very confident in their predictions.

\begin{table}[t]
    \centering
    \tablestyle{3pt}{1}
\begin{tabular}{ccccccccc}
\toprule
& \multicolumn{2}{c}{\wavvecsize{base} 100h} & \multicolumn{2}{c}{\wavvecsize{base} 960h} & \multicolumn{2}{c}{\wavvecsize{large} 100h} & \multicolumn{2}{c}{\wavvecsize{large} 960h} \\
\cmidrule(l{2pt}r{2pt}){2-3}
\cmidrule(l{2pt}r{2pt}){4-5}
\cmidrule(l{2pt}r{2pt}){6-7}
\cmidrule(l{2pt}r{2pt}){8-9}
& Infer. Time & WER (\%) & Infer. Time & WER (\%) & Infer. Time & WER (\%) & Infer. Time & WER (\%) \\
\midrule
Best path & 30.8 & 13.5687 & 30.7 & 8.8777 & 72.9 & 9.2663 & 72.3 & 6.5302 \\
Viterbi & 31.8 & 13.5687 & 32.1 & 8.8777 & 73.7 & 9.2663 & 73.1 & 6.5302 \\
\bottomrule
\end{tabular}
\vspace{3pt}
    \caption{Best path decoding vs. Viterbi decoding using official \wavvec ASR models. We show a higher precision of WER to emphasize performance is identical.}
    \label{tab:greedy_vs_viterbi}
    \tabbelowsqueeze
\end{table}

\paragraphsq{Inference Time with Language Models}

\shortautoref{tab:lm_decode} shows \wavvec's inference time with or without LM.
Decoding with LM improves the WER significantly, but inference time is also increased dramatically.
It is likely that the beam search implementation is sequential and CPU-bound which slows down decoding.
Reducing the inference time with LM is an important direction for future work.
Alternatively, recent work~\citep{xu2020iterative,Xu2020SelftrainingAP} shows pseudo-labeling can improve the model performance and close the gap between with and without LM. This can also solve the slow inference time with LM. 
We use \wavvec's official inference script in these experiments. The slowdown depends on the CPU type. 
We observe smaller inference time overhead when decoding on faster CPUs, but for consistency, we use the same type of hardware as our pre-training setup.

\begin{table}[t]
    \centering
    \tablestyle{3pt}{1}
    \begin{tabular}{ccccccc}
    \toprule
    & \multicolumn{3}{c}{Inference Time (s)} & \multicolumn{3}{c}{WER (\%)} \\
    \cmidrule(l{2pt}r{2pt}){2-4}
    \cmidrule(l{2pt}r{2pt}){5-7}
    & No LM & LM (beam=5) & LM (beam=50) & No LM & LM (beam=5) & LM (beam=50) \\
    \midrule
    \wavvecsize{tiny} 100h (100K) & 7.5\mypm{0.04} & 20.1\mypm{0.10} & 119.0\mypm{0.81} & 40.7 & 31.8 & 23.4 \\
    \wavvecsize{base} 100h (400K) & 30.8\mypm{0.05} & 42.7\mypm{0.28} & 128.5\mypm{0.82} & 13.6 & 10.7 & 8.5 \\
    \bottomrule
    \end{tabular}
    \vspace{3pt}
    \caption{\wavvec's inference time and WER with or without an LM. Decoding with an LM increases the inference time significantly.}
    \label{tab:lm_decode}
    \tabbelowsqueeze
\end{table}
\section{Additional Experiments on the Kernel size of Downsampling}\label{app:sec:k127}

While in \shortautoref{subsec:wfe_vs_context} we show that with small inferenc budget reducing the kernel size of the downsampling layer allows us to increase the size of WFE and leads to better performance. We conduct additional experiments with various model size attempting to understand why \citet{baevski2020wav2vec2} chose a large convolutional kernel size for their models. \shortautoref{tab:ls_100h_100k_k127} shows the performance of various models with kernel size 127. As we can see, at small model regime the overhead contributes to a large portion of the inference time and is prohibitive, while for large models, this overhead becomes relatively small and the boost of performance makes it favorable. \shortautoref{tab:ls_100h_400k_k127} shows the performance of the model pre-trained for 400K updates and using kernel size 127 leads to better WERs especially on dev-other and test-other sets.
\shortautoref{tab:ls_1m_1h_10h_k127} shows the performance with less supervision.
As we can see, using kernel size 127 leads to better WERs when decoding with a language model.

\begin{table}[h]
\centering
\tablestyle{4pt}{1.1}
\begin{tabular}{lrcccccc}
\toprule
& & \multicolumn{2}{c}{Time} & \multicolumn{4}{c}{WER (No LM / 4-gram LM beam=50)} \\
\cmidrule(l{2pt}r{2pt}){3-4}
\cmidrule(l{2pt}r{2pt}){5-8}
Model & \# Param. & PT (h) & Infer. (s) & dev-clean & dev-other & test-clean & test-other \\
\midrule
\finalmodelDsize{tiny}  & 24.1 & 23.8 & 7.5\mypm{0.02} & 10.1 / 4.4 & 22.3 / 13.4 & 10.4 / 4.9 & 22.8 / 13.9 \\
\finalmodelDsize{tiny} (k127) & 25.0 & 24.7 & 8.5\mypm{0.03}  & 10.0 / 4.3 & 22.1 / 13.3 & 10.4 / 4.9 & 22.9 / 13.6 \\
\midrule
\finalmodelDsize{small} & 41.0 & 38.1 & 9.6\mypm{0.04} & 7.5 / 3.6 & 17.9 / 11.1 & 7.8 / 4.2 & 18.2 / 11.4 \\
\finalmodelDsize{small} (k127) & 42.6 & 39.9 & 10.9\mypm{0.03} & 7.8 / 3.7 & 18.1 / 11.0 & 7.9 / 4.2 & 18.4 / 11.6 \\
\midrule

\finalmodelDsize{mid} & 78.8 & 51.7 & 16.5\mypm{0.03} & 6.3 / 3.2 & 14.0 / 9.3 & 6.4 / 3.8 & 14.2 / 9.5 \\
\finalmodelDsize{mid} (k127) & 80.4 & 56.6 & 17.9\mypm{0.05} & 6.1 / 3.1 & 13.5 / 8.9 & 6.3 / 3.7 & 13.8 / 9.4 \\
\bottomrule

\end{tabular}
\vspace{+3pt}
\caption{Comparing kernel sizes 31 and 127 on Libri-Speech 100h-960h semi-supervised setup pretrained for 100K updates.}
\label{tab:ls_100h_100k_k127}
\tabbelowsqueeze
\end{table}
\begin{table}[h]
\centering
\tablestyle{4pt}{1.1}
\begin{tabular}{lrcccccc}
\toprule
& & \multicolumn{2}{c}{Time} & \multicolumn{4}{c}{WER (No LM / 4-gram LM beam=50)} \\

\cmidrule(l{2pt}r{2pt}){3-4}
\cmidrule(l{2pt}r{2pt}){5-8}
Model & \# Param. & PT (GPU-day) & Infer. (s) & dev-clean & dev-other & test-clean & test-other \\
\midrule
\wavvecsize{base} & 94.4 & 102.4(73.5) & 30.8\mypm{0.05} & 6.1 / 2.9 & 13.6 / 8.5 & 6.1 / 3.5 & 13.3 / 8.7 \\
\midrule
\finalmodelDsize{mid} & 78.8 & 68.9 & 16.5\mypm{0.03} & 4.8 / 2.7 & 11.1 / 7.6 & 4.9 / 3.3 & 11.5 / 8.2 \\

\finalmodelDsize{mid} (k127) & 80.4 & 75.4 & 17.9\mypm{0.05} & 5.0 / 2.7 & 10.8 / 7.5 & 5.0 / 3.2 & 10.9 / 7.9 \\

\bottomrule
\end{tabular}
\vspace{+3pt}
\caption{Comparing kernel sizes 31 and 127 on LibriSpeech 100h-960h semi-supervised setup pretrained for 400K updates.}
\label{tab:ls_100h_400k_k127}
\tabbelowsqueeze
\end{table}
\begin{table}[h]
\centering
\tablestyle{4pt}{1.1}
\begin{tabular}{lccccccc}
\toprule
& Inference & \multicolumn{2}{c}{10m} & \multicolumn{2}{c}{1h} & \multicolumn{2}{c}{10h} \\
\cmidrule(l{2pt}r{2pt}){3-4}
\cmidrule(l{2pt}r{2pt}){5-6}
\cmidrule(l{2pt}r{2pt}){7-8}
Model & Time (s) & test-clean & test-other & test-clean & test-other & test-clean & test-other \\
\midrule
\wavvecsize{base} & 30.8\mypm{0.05} & 47.8\mypm{1.16} & 54.0\mypm{1.18} & 33.3\mypm{2.73} & 40.0\mypm{2.34} & 13.3\mypm{0.27} & 22.6\mypm{0.15} \\
 + LM & & 19.7\mypm{0.26} & 28.2\mypm{0.23} & 9.9\mypm{0.48} & 18.4\mypm{0.65} & 6.4\mypm{0.08} & 13.8\mypm{0.13} \\
\midrule
\finalmodelDsize{mid} & 16.5\mypm{0.03} & 54.2\mypm{1.36} & 59.5\mypm{1.09} & 30.4\mypm{0.08} & 36.8\mypm{0.18} & 12.3\mypm{0.06} & 20.7\mypm{0.07} \\
+ LM & & 25.5\mypm{1.06} & 33.4\mypm{1.20} & 9.6\mypm{0.08} & 17.5\mypm{0.11} & 5.9\mypm{0.04} & 13.0\mypm{0.07} \\
\midrule
\finalmodelDsize{mid} (k=127) & 17.9\mypm{0.05} & 48.4\mypm{0.85} & 52.7\mypm{0.90} & 32.8\mypm{0.14} & 37.6\mypm{0.28} & 11.6\mypm{0.11} & 19.3\mypm{0.01} \\
 + LM & & \textbf{18.2}\mypm{0.18} & \textbf{25.8}\mypm{0.26} & 9.6\mypm{0.06} & \textbf{16.6}\mypm{0.11} & 5.6\mypm{0.05} & 12.4\mypm{0.06} \\
\midrule
\finalmodelDsize{base+} & 27.8\mypm{0.05} & \textbf{43.4}\mypm{0.54} & \textbf{49.3}\mypm{0.42} & \textbf{21.5}\mypm{0.22} & \textbf{28.9}\mypm{0.11} & \textbf{11.0}\mypm{0.35} & \textbf{18.1}\mypm{0.06} \\
+ LM & & 22.6\mypm{0.40} & 30.6\mypm{0.25} & \textbf{9.5}\mypm{0.37} & 17.4\mypm{0.38} & \textbf{5.2}\mypm{0.06} & \textbf{11.7}\mypm{0.24} \\
\bottomrule
\end{tabular}
\vspace{+3pt}
\caption{LibriSpeech results with 10m, 1h, or 10h supervised data. All the models are pre-trained for 100K updates on LibriSpeech 960h. We report inference times on dev-clean and WERs without LM and with 4-gram LM (beam size 50). The mean and standard deviations are calculated from three random runs. Similar to BERT-large~\citep{phang2018sentence}, we observe that our \finalmodelDsize{base+} is unstable during fine-tuning. We run five random runs and stop two degenerate runs after 3K updates.}
\label{tab:ls_1m_1h_10h_k127}
\tabbelowsqueeze
\end{table}
\section{Additional Transferring Results}\label{sec:additional_transfer_results}

\shortautoref{tab:transfer_results} shows additional results of transferring LibriSpeech pre-trained models to three out-of-domain datasets. We share the performance of the models with trained with only 100K updates enable future works for quick comparison.

\begin{table}[h]
\centering
\tablestyle{3pt}{1.1}
\scriptsize
\begin{tabular}{clccccccccc}
\toprule
& & \multicolumn{3}{c}{TED-LIUM 3 (10h)} & \multicolumn{3}{c}{VoxPopuli (10h)} & \multicolumn{3}{c}{Fisher+Switchboard (10h)} \\
\cmidrule(l{2pt}r{2pt}){3-5}
\cmidrule(l{2pt}r{2pt}){6-8}
\cmidrule(l{2pt}r{2pt}){9-11}
PT Iter. & Model & Time (s) & dev (\%) & test (\%) & Time (s) & dev (\%) & test (\%) & Time (s) & dev (\%) & test (\%) \\
\midrule
\multirow{6}{*}{ 100K } & W2V2-base & 10.2\mypm{0.03} & 21.2\mypm{0.47} & 19.7\mypm{0.38} & 30.8\mypm{0.22} & 22.5\mypm{0.08} & 23.3\mypm{0.20} & 33.8\mypm{0.12} & 28.8\mypm{0.06} & 28.9\mypm{0.15} \\
& + LM & & 11.3\mypm{0.21} & 11.6\mypm{0.04} & & 13.2\mypm{0.06} & 13.4\mypm{0.16} & & 19.8\mypm{0.12} & 20.2\mypm{0.15} \\
\cmidrule{2-11}
& \finalmodelDsize{mid} & 6.9\mypm{0.02} & 18.4\mypm{0.12} & 18.0\mypm{0.07} & 20.3\mypm{0.08} & 20.9\mypm{0.08} & 21.6\mypm{0.07} & 21.6\mypm{0.07} & 27.6\mypm{0.15} & 28.1\mypm{0.06} \\
& + LM & & 10.3\mypm{0.02} & 11.3\mypm{0.07} & & 12.6\mypm{0.10} & 13.1\mypm{0.10} & & 19.9\mypm{0.10} & 20.4\mypm{0.12} \\
\cmidrule{2-11}
& \finalmodelDsize{mid}-k127 & 7.0\mypm{0.07} & 18.1\mypm{0.25} & 17.1\mypm{0.08} & 20.8\mypm{0.09} & 19.7\mypm{0.05} & 20.2\mypm{0.11} & 25.0\mypm{0.04} & 26.9\mypm{0.06} & 27.2\mypm{0.12} \\
& + LM & & 10.2\mypm{0.24} & 10.4\mypm{0.13} & & 12.2\mypm{0.07} & 12.6\mypm{0.07} & & 18.7\mypm{0.10} & 19.0\mypm{0.06} \\
\cmidrule{2-11}
& \finalmodelDsize{base+} & 11.2\mypm{0.01} & 15.7\mypm{0.05} & 15.6\mypm{0.02} & 33.8\mypm{0.04} & 17.8\mypm{0.08} & 18.1\mypm{0.12} & 36.8\mypm{0.03} & 25.3\mypm{0.62} & 25.9\mypm{0.59} \\
& + LM & & 9.7\mypm{0.09} & 10.3\mypm{0.03} & & 11.7\mypm{0.12} & 11.9\mypm{0.03} & & 18.7\mypm{0.44} & 19.1\mypm{0.30} \\
\midrule
\multirow{6}{*}{ 400K } & \wavvecsize{base} & 10.2\mypm{0.03} & 15.7\mypm{0.08} & 14.8\mypm{0.13} & 30.8\mypm{0.22} & 17.5\mypm{0.10} & 18.1\mypm{0.03} & 33.8\mypm{0.12} & 23.2\mypm{0.06} & 23.2\mypm{0.10} \\
& + LM & & 9.2\mypm{0.09} & 9.3\mypm{0.09} & & 11.2\mypm{0.07} & 11.6\mypm{0.01} & & 16.5\mypm{0.06} & 16.7\mypm{0.00} \\
\cmidrule{2-11}
& \finalmodelDsize{mid} & 6.9\mypm{0.02} & 14.4\mypm{0.11} & 13.9\mypm{0.04} & 20.3\mypm{0.08} & 17.5\mypm{0.10} & 18.1\mypm{0.03} & 21.6\mypm{0.07} & 24.1\mypm{0.67} & 24.3\mypm{0.49} \\
& + LM & & 9.0\mypm{0.24} & 9.2\mypm{0.11} & & 10.9\mypm{0.22} & 11.2\mypm{0.26} & & 18.0\mypm{0.44} & 18.3\mypm{0.53} \\
\cmidrule{2-11}
& \finalmodelDsize{mid}-k127 & 7.0\mypm{0.07} & 14.7\mypm{0.41} & 14.1\mypm{0.44} & 20.8\mypm{0.09} & 16.4\mypm{0.17} & 16.8\mypm{0.27} & 25.0\mypm{0.04} & 23.4\mypm{0.21} & 23.8\mypm{0.21} \\
& + LM & & 8.5\mypm{0.12} & 9.0\mypm{0.16} & & 10.7\mypm{0.10} & 11.0\mypm{0.07} & & 16.5\mypm{0.07} & 16.8\mypm{0.35} \\
\cmidrule{2-11}
& \finalmodelDsize{base+} & 11.2\mypm{0.01} & 12.8\mypm{0.28} & 12.2\mypm{0.14} & 33.8\mypm{0.04} & 14.8\mypm{0.27} & 15.5\mypm{0.35} & 36.8\mypm{0.03} & 20.5\mypm{0.10} & 20.5\mypm{0.06} \\
& + LM & & 8.3\mypm{0.39} & 8.7\mypm{0.22} & & 10.3\mypm{0.13} & 10.9\mypm{0.39} & & 15.6\mypm{0.21} & 15.7\mypm{0.06} \\
\bottomrule
\end{tabular}
\vspace{+3pt}
\caption{Inference time and WER of LibriSpeech pre-trained models transferred to TED-LIUM 3, VoxPopuli, and Fisher+Switchboard datasets with only 10h labels. \finalmodel-D mid outperforms W2V2 base on all settings while being at least 30\% faster. We report the inference time on the dev sets. The mean and standard deviations are computed over three random runs.}
\label{tab:transfer_results}
\tabbelowsqueeze
\end{table}

\section{Disentangled Attention} \label{app:sec:disent_attn}
\citet{He2020DeBERTaDB} introduced disentangled attention as a component of their DeBERTa model.
Unlike Transformer which adds absolute positional embeddings to the content embeddings at the beginning,
disentangled attention keeps the positional embeddings and content embeddings separate and has three attention components in its attention weight computation: (a) content-to-content, (b) content-to-position, and (c) position-to-content. 
Given the content embeddings $\mC \in \R^{T \times d}$ and position embeddings $\mP \in \R^{(2k + 1) \times d}$, where k is the maximum relative position, the output embeddings $\mO \in \R^{T \times d}$ are:

\begin{small}
\begin{equation}
    \mO = \mathrm{softmax} \left( \frac{\tilde{\mA}}{\sqrt{3d}} \right) \mV^c,
    \quad
    \tilde{\mA}_{i, j} = 
    \underbrace{\left( \mQ^c\mK^{c\top} \right)_{i, j}}_\text{(a) content-to-content} 
    + \underbrace{\left( \mQ^c \mK^{p\top} \right)_{i, \delta(i, j)}}_\text{(b) content-to-position}
    + \underbrace{\left( \mQ^p \mK^{c\top} \right)_{\delta(i, j), j}}_\text{(c) position-to-content}\;\;,
\end{equation}
\end{small}

\noindent
where $\mQ^c = \mC \mW^{q,c}$, $\mK^c = \mC \mW^{k,c}$, $\mV^c = \mC \mW^{v,c}$, $\mQ^p = \mP \mW^{q, p}$, $\mK^p = \mP \mW^{k, p}$, and $\mW^{q,c}, \mW^{k,c}, \mW^{v,c}, \mW^{q,p}, \mW^{k,p} \in \R^{d \times d}$ are trainable projection weights.
$\delta(i, j)$ is the relative position of $i, j$ clamped to $[-k, k]$ and is used to index the corresponding row of $\mP$ or its products.

Unlike conventional self-attention which only has content-to-content and four matrix multiplication operations in total, disentangled attention has nine multiplication operations in total and is much slower to compute than the conventional self-attention.

\end{document}